\documentclass[9pt,technote]{IEEEtran}

\usepackage{makeidx}
\usepackage[bookmarks=true]{hyperref}

\usepackage{amssymb,amsmath,amsthm}
\usepackage{graphicx}
\usepackage{setspace}
\usepackage{mathrsfs}
\usepackage{enumerate}
\usepackage{multirow}

\newcommand{\cP}{\mathcal{P}}

\newcommand{\taumin}{\tau^{\min}}
\newcommand{\taumax}{\tau^{\max}}

\newcommand{\bfa}{\mathbf{a}}
\newcommand{\bfb}{\mathbf{b}}
\newcommand{\bfc}{\mathbf{c}}

\newcommand{\bfg}{\mathbf{g}}

\newcommand{\bfq}{\mathbf{q}}
\newcommand{\bff}{\mathbf{f}}

\newcommand{\bfA}{\mathbf{A}}
\newcommand{\bfB}{\mathbf{B}}
\newcommand{\bfC}{\mathbf{C}}

\newcommand{\bfM}{\mathbf{M}}

\newcommand{\ud}{\mathrm{d}}
\newcommand{\bftau}{\boldsymbol{\mathbf{\tau}}}

\newcommand{\beg}{\mathrm{beg}}
\newcommand{\fin}{\mathrm{end}}

\newcommand{\MVC}{\mathrm{MVC}}

\newtheorem{Prop}{Proposition}
\newtheorem*{Ex*}{Examples}

\title{A General, Fast, and Robust Implementation of the Time-Optimal
  Path Parameterization Algorithm}

\author{Quang-Cuong Pham\\ 
  School of Mechanical and Aerospace
  Engineering\\ 
  Nanyang Technological University, Singapore}

\begin{document}

\maketitle
\thispagestyle{empty}
\pagestyle{empty}

\begin{abstract}
  Finding the Time-Optimal Parameterization of a given Path (TOPP)
  subject to kinodynamic constraints is an essential component in many
  robotic theories and applications. The objective of this article is
  to provide a general, fast and robust implementation of this
  component. For this, we give a complete solution to the issue of
  dynamic singularities, which are the main cause of failure in
  existing implementations. We then present an open-source
  implementation of the algorithm in C++/Python and demonstrate its
  robustness and speed in various robotics settings.
\end{abstract}

\section{Introduction}
\label{sec:intro}

Time-optimal motion planning plays a key role in many areas of
robotics and automation, from industrial to mobile, to service
robotics. While the problem of (optimal) \emph{path planning under
  geometric constraints} can be considered as essentially solved in
both theory and in practice~(see e.g. \cite{Lav06book}), general and
efficient solutions to the (optimal) \emph{trajectory planning under
  kinodynamic\,\footnote{Geometric constraints -- such as joints
    limits or obstacle avoidance -- depend only on the configuration
    of the robot, while kinodynamic constraints -- such as bounds on
    joint velocity, acceleration and torque, or dynamic balance --
    involve also higher-order time derivatives of the robot
    configuration.}  constraints}~\cite{DonX93acm} are still
lacking. We argue that Time-Optimal Path
Parameterization\,\footnote{\emph{Parameterizing} a given geometric
  path consists in finding a time law to traverse the path, thereby
  transforming it into a \emph{trajectory}.  \emph{Time-optimal}
  parameterization seeks to minimize the traversal time under given
  kinodynamic constraints.} (TOPP) may constitute an efficient tool to
address the latter problem.

There are at least three types of kinodynamic motion planning problems
where TOPP is useful or even indispensable. First, some applications
such as painting or welding require specifically tracking a predefined
path. Second, even when there is no \emph{a priori} necessity to track
a predefined path, it can be efficient to \emph{decouple} the optimal
trajectory planning problem into two simpler, more tractable
sub-problems: (i) generate a set of paths in the robot configuration
space, (ii) optimally time-parameterize these paths and pick the path
with the fastest optimal
parameterization~\cite{Bob88jra,SD91tra}. Third, it was recently
suggested that TOPP can also be used to address the \emph{feasibility}
problem~\cite{PhaX13rss}, i.e. finding \emph{one} feasible trajectory
in a challenging context, as opposed to selecting an optimal trajectory
in a context where it is relatively easy to find many feasible
trajectories.

Since the path is
constrained, the only ``degree of freedom'' to optimize is the scalar
function $t\mapsto s(t)$ which represents the ``position'' on the path
at each time instant. If the system dynamics and constraints are of
second-order, one can next search for the optimal function in the
2-dimensional space $(s,\dot s)$. There are basically three families
of methods to do so.

\emph{Dynamic programming} -- The first family of methods divide the
$(s,\dot s)$ plane into a grid and subsequently uses a dynamic
programming approach to find the optimal trajectory in the $(s,\dot
s)$ plane~\cite{SM86tac}.

\emph{Convex optimization} -- The second family of methods discretize
only the $s$-axis (into $N$ segments) and subsequently convert the
original problem into a convex optimization problem in $O(N)$
variables and $O(N)$ equality and inequality
constraints~\cite{VerX09tac,Hau13rss,DebX13tr}. These methods have the
advantage of being versatile (they can for instance trade off time
duration with other objectives such as energy or torque rate) and can
rely on existing efficient convex optimization packages.

\emph{Numerical integration} -- The third family of methods are based
on the Pontryagin Maximum Principle: the optimal trajectory in the
$(s,\dot s)$ plane is known to be ``bang-bang'' and can thus be found
by integrating successively the maximum and minimum
accelerations~$\ddot s$, see Section~\ref{sec:overview} for
details. This approach is theoretically faster than the previous two
since it exploits the bang-bang structure of the optimization problem
(and we shall show that it is indeed faster in practice). However, to
our knowledge, there is no available general and efficient
implementation, perhaps because of the programming difficulties
involved by this approach and of the robustness issues associated with
the so-called \emph{dynamic
  singularities}~\cite{SL92jdsmc,KS12rss,Pha13iros}, see details in
Section~\ref{sec:robust}.

Note that all three families can be applied to a wide variety of robot
dynamics and constraints, such as manipulators subject to torque
bounds~\cite{BobX85ijrr}, humanoids subject to joint velocity and
accelerations bounds~\cite{KS12rss,Hau13rss}, mobile robots or
humanoids subject to balance and friction
constraints~\cite{PN12humanoids,Hau14icra}, non-holonomic
robots~\cite{BL01tac}, etc.

The goal of this article is to provide a \emph{general}, \emph{fast}
and \emph{robust} implementation of TOPP. For this, we follow the
theoretically faster numerical integration approach.  To make it
robust, we address the aforementioned critical issue of \emph{dynamic
  singularities}. Such singularities arise in a large proportion of
real-world problem instances of TOPP, and are one of the main causes
of failure of existing implementations of the numerical integration
approach. Note that dynamic singularities also cause jitters in the
convex optimization approach (see e.g. Fig. 4 of~\cite{VerX09tac}) but
are probably not as critical there as in the numerical integration
approach. Yet, in most works devoted to the TOPP algorithm, from
original articles~\cite{PJ87jra,SL92jdsmc,SY89tra,KS12rss} to
reference textbooks~\cite{ChoX05book}, the characterization and
treatment of these singularities were not done completely
correctly. In Section~\ref{sec:robust}, we derive a complete
characterization of dynamic singularities and suggest how to
appropriately address these singularities. The development extends our
previous contribution in the case of torque bounds (presented at IROS
2013 \cite{Pha13iros}) to the general case. In
Section~\ref{sec:implementation}, we present an open-source
implementation in C++/Python. We show that our implementation is
robust and about one order of magnitude faster than existing
implementations of the convex optimization
approach~\cite{VerX09tac,DebX13tr,Hau13rss}. This improvement is
particularly crucial for the global trajectory optimization problem or
the feasibility problem mentioned earlier, which require calling the
TOPP routine tens or hundreds of thousands times. Finally
Section~\ref{sec:conclusion} concludes by briefly discussing the
obtained results and future research directions.

\section{Improving the robustness of the\\numerical integration approach}
\label{sec:robust}

\subsection{General formulation of the TOPP problem}
\label{sec:general}

Let $\bfq$ be a $n$-dimensional vector representing the configuration
of a robot system. Consider second-order inequality constraints of the
form
\begin{equation}
  \label{eq:gen}
  \bfA(\bfq)\ddot\bfq + \dot\bfq^\top \bfB(\bfq) \dot\bfq + \bff(\bfq) \leq 0, 
\end{equation}
where $\bfA(\bfq)$, $\bfB(\bfq)$ and $\bff(\bfq)$ are respectively an
$M\times n$ matrix, an $n\times M \times n$ tensor and an
$M$-dimensional vector.

Note that ``direct'' velocity bounds of the form
\begin{equation}
  \label{eq:velo}
\dot\bfq^\top \bfB_v(\bfq) \dot\bfq + \bff_v(\bfq) \leq 0,
\end{equation}
can also be taken into account, see footnote~\ref{foot:zla}
and~\cite{Zla96icra,LL98er}.

Inequality~(\ref{eq:gen}) is very general and may represent a large
variety of second-order systems and constraints. As an example,
consider an $n$-dof manipulator with dynamics
  \begin{equation}
    \label{eq:manip}
    \bfM(\bfq)\ddot\bfq+\dot\bfq^\top\bfC(\bfq)\dot\bfq+\bfg(\bfq)=\bftau.
  \end{equation}
Assume that the manipulator is subject to lower and upper bounds on
the joint torques, that is, for any joint $i$ and time $t$,
  \[
  \taumin_i \leq \tau_i(t) \leq \taumax_i.
  \]
  Clearly, these torque bounds can be put in the form of~(\ref{eq:gen})
  with
  \[
  \tiny
  \bfA(\bfq)=\left(\begin{array}{r}
      \bfM(\bfq)\\
      -\bfM(\bfq)
  \end{array}
  \right),
  \
    \bfB(\bfq)=\left(\begin{array}{r}
      \bfC(\bfq)\\
      -\bfC(\bfq)
  \end{array}
  \right),
  \
  \bff(\bfq)=\left(\begin{array}{r}
      \bfg(\bfq)-\bftau^{\max}\\
      -\bfg(\bfq)+\bftau^{\min}
  \end{array}
  \right),
  \]
  where $\bftau^{\max} = (\tau_1^{\max},\dots,\tau_n^{\max})^\top$ and
  $\bftau^{\min} = (\tau_1^{\min},\dots,\tau_n^{\min})^\top$.

  Consider now a path $\cP$ -- represented as the underlying path of a
  trajectory $\bfq(s)_{s\in[0,s_\fin]}$ -- in the configuration space.
  Assume that $\bfq(s)_{s\in[0,s_\fin]}$ is $C^1$- and piecewise
  $C^2$-continuous (note that how to generate such smooth initial
  trajectories, especially for closed kinematic chains, is an
  interesting problem on its own). We are interested in
  \emph{time-parameterizations} of $\cP$ -- or
  time-\emph{re}parameterizations of $\bfq(s)_{s\in[0,s_\fin]}$ --
  which are increasing \emph{scalar functions} $s : [0,T']\rightarrow
  [0,s_\fin]$. Differentiating $\bfq(s(t))$ with respect to $t$ yields
\begin{equation}
  \label{eq:dotq}
  \dot\bfq= \bfq_s\dot s, \quad \ddot\bfq= \bfq_s  \ddot s + \bfq_{ss}
  \dot s^2 ,
\end{equation}
where dots denote differentiations with respect to the time parameter
$t$ and $\bfq_s=\frac{\ud \bfq}{\ud s}$ and $\bfq_{ss}=\frac{\ud^2
  \bfq}{\ud s^2}$.
Substituting (\ref{eq:dotq}) into (\ref{eq:gen}) then leads to
\[
\ddot s \bfA(\bfq)\bfq_s + \dot s^2 \bfA(\bfq)\bfq_{ss}+\dot s^2
\bfq_s^\top\bfB(\bfq)\bfq_s + \bff(\bfq) \leq 0,
\]
which can be rewritten as
\begin{equation}
  \label{eq:gen2}
  \ddot s \bfa(s) + \dot s^2 \bfb(s) + \bfc(s) \leq 0,\quad\textrm{where}  
\end{equation}
\begin{eqnarray}
  \label{eq:toto}
  \bfa(s)&=&\bfA(\bfq(s))\bfq_s(s),\nonumber\\
  \bfb(s)&=&\bfA(\bfq(s))\bfq_{ss}(s) + \bfq_s(s)^\top\bfB(\bfq(s))\bfq_s(s),\\
  \bfc(s)&=&\bff(\bfq(s)).\nonumber 
\end{eqnarray}

Equation~(\ref{eq:gen2}) constitutes another level of abstraction,
which is particularly convenient for computer implementation: it
suffices indeed to evaluate the $M$-dimensional vectors $\bfa$, $\bfb$
and $\bfc$ along the path and feed these vectors as inputs to the
optimization algorithm. From this formulation, one can also remark
that it is \emph{not necessary} to evaluate the full matrices $\bfA$
and tensors $\bfB$ (which are of sizes $M\times n$ and $n\times
M\times n$), but only their \emph{products} (of sizes $M$) with
$\bfq_s$ and $\bfq_{ss}$. In the case of torque bounds, the Recursive
Newton-Euler method for inverse dynamics~\cite{WO82jdsmc} allows
computing these products without ever evaluating the full $\bfA$ and
$\bfB$. Finally, this formulation allows very easily \emph{combining}
different types of constraints: for each $s$, it suffices to
concatenate the vectors $\bfa(s)$ corresponding to the different
constraints, and similarly for the vectors $\bfb(s)$ and $\bfc(s)$.

\subsection{Review of the numerical integration approach}
\label{sec:overview}

Each row $i$ of equation~(\ref{eq:gen2}) is of the form
\[
a_i(s)\ddot s + b_i(s)\dot s^2 + c_i(s) \leq 0.
\]
There are three cases:
\begin{itemize}
\item if $a_i(s)>0$, then one has $\ddot s \leq
  \frac{-c_i(s)-b_i(s)\dot s^2}{a_i(s)}$.  Define the \emph{upper
    bound} $\beta_i=\frac{-c_i(s)-b_i(s)\dot s^2}{a_i(s)}$;
\item if $a_i(s)<0$, then one has $\ddot s \geq
  \frac{-c_i(s)-b_i(s)\dot s^2}{a_i(s)}$. Define the \emph{lower
    bound} $\alpha_i=\frac{-c_i(s)-b_i(s)\dot s^2}{a_i(s)}$;
\item if $a_i(s)=0$, then $s$ is a
  ``\emph{zero-inertia}'' point~\cite{PJ87jra,SL92jdsmc}.
\end{itemize}

One then has a certain number of $\alpha_p$ and $\beta_q$. Their total
number is $\leq M$, with equality when $s$ is not a zero-inertia
point.  One can next define for each $(s,\dot s)$
\begin{equation}
  \label{eq:tata}
  \alpha(s,\dot s)=\max_p \alpha_p(s,\dot s),\nonumber \quad
  \beta(s,\dot s)=\min_q \beta_q(s,\dot s).\nonumber 
\end{equation}

From the above transformations, one can conclude that
$\bfq(s(t))_{t\in[0,T']}$ satisfies the constraints~(\ref{eq:gen}) if
and only if
\begin{equation}
\label{eq:bounds}
\forall t\in[0,T']\quad \alpha(s(t),\dot s(t)) \leq \ddot s(t) \leq
\beta(s(t),\dot s(t)).  
\end{equation}

Note that $(s,\dot s)\mapsto(\dot s,\alpha(s,\dot s))$ and $(s,\dot
s)\mapsto(\dot s,\beta(s,\dot s))$ can be viewed as two vector fields
in the $(s,\dot s)$ plane. One can integrate velocity profiles
following the field $(\dot s,\alpha(s,\dot s))$ (from now on, $\alpha$
in short) to obtain \emph{minimum acceleration} profiles (or
$\alpha$-profiles), or following the field $\beta$ to obtain
\emph{maximum acceleration} profiles (or $\beta$-profiles).

Next, observe that if $\alpha(s,\dot s)>\beta(s,\dot s)$ then, from
(\ref{eq:bounds}), there is no possible value for $\ddot s$. Thus, to
be valid, every velocity profile must stay below the maximum velocity
curve\,\footnote{\label{foot:zla}If ``direct'' velocity bounds such as
  in equation~(\ref{eq:velo}) are considered, then they induce another
  maximum velocity curve, noted $\MVC_\mathrm{direct}$. In this case,
  every velocity profile must stay below
  $\min(\MVC,\MVC_\mathrm{direct})$. The treatment of
  $\MVC_\mathrm{direct}$ was initiated in~\cite{Zla96icra,LL98er} and
  completed and implemented by us, see
  \url{https://github.com/quangounet/TOPP/releases}.} ($\MVC$ in
short) defined by
\[ 
\MVC(s)=\left\{
  \begin{array}{cl}
    \min \{\dot s\geq 0: \alpha(s,\dot s)= \beta(s,\dot
    s)\}&\mathrm{if}\ \alpha(s,0) \leq \beta(s,0),\\
    0&\mathrm{if}\ \alpha(s,0) > \beta(s,0).
  \end{array}
\right.
\]

It was shown (see e.g. \cite{SL92jdsmc}) that the time-minimal
velocity profile is obtained by a \emph{bang-bang}-type control, i.e.,
whereby the optimal profile follows alternatively the $\beta$ and
$\alpha$ fields while always staying below the $\MVC$. More precisely,
the algorithm to find the time-optimal parameterization of $\cP$
starting and ending with the desired linear velocities $v_\beg$ and
$v_\fin$ is as follows (see Fig.~\ref{fig:bobrow}):

\begin{figure}[htp]
    \centering
    \includegraphics[width=7cm]{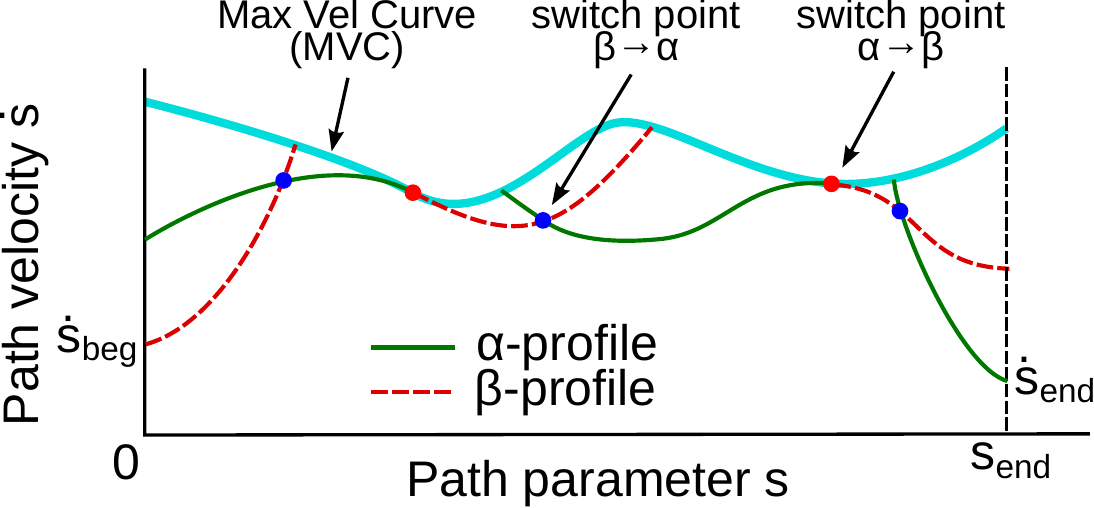}
    \caption{MVC and $\alpha$-, $\beta$-profiles in the numerical
      integration approach.}
  \label{fig:bobrow}
\end{figure}

\noindent\textbf{1.} In the $(s,\dot s)$ plane, start from ($s=0,\dot
  s=v_\beg/\|\bfq_s(0)\|$) and integrate forward following $\beta$ until
  hitting either
  \begin{enumerate}[(i)]
  \item the $\MVC$, in this case go to step 2;
  \item the horizontal line $\dot s=0$, in this case  the
    path is not dynamically traversable;
  \item the vertical line $s=s_\fin$, in this case go to step~3.
  \end{enumerate}

\noindent\textbf{2.} Search forward along the $\MVC$ for the next candidate
  $\alpha\to\beta$ switch point (cf. Section~\ref{sec:charact}). From
  such a switch point:
  \begin{enumerate}[(a)]
  \item integrate \emph{backward} following $\alpha$, until
    \emph{intersecting} a forward $\beta$-profile (from step 1 or
    recursively from the current step 2). The intersection point
    constitutes a $\beta\to\alpha$ switch point;
  \item integrate \emph{forward} following $\beta$. Then continue as
    in step 1.
\end{enumerate}
The resulting forward profile will be the concatenation of the
intersected forward $\beta$-profile, the backward $\alpha$-profile
obtained in (a), and the forward $\beta$-profile obtained in~(b).

\noindent\textbf{3.} Start from ($s=s_\fin,\dot s=v_\fin/\|\bfq_s(s_\fin)\|$)
  and integrate \emph{backward} following $\alpha$, until intersecting
  a forward profile obtained in steps 1 or 2.  The intersection point
  constitutes a $\beta\to\alpha$ switch point. The final profile will
  be the concatenation of the intersected forward profile and the
  backward $\alpha$-profile just computed.

  From the above presentation, it appears that finding the
  $\alpha\to\beta$ switch points is crucial for the numerical
  integration approach. It was shown in
  \cite{PJ87jra,SY89tra,SL92jdsmc} that a given point $s$ is a
  $\alpha\to\beta$ switch point only in the following three cases:
\begin{itemize}
\item the $\MVC$ is \emph{discontinuous} at $s$. In this case $s$ is
  labeled as a ``discontinuous'' switch point;
\item the $\MVC$ is \emph{continuous} but \emph{undifferentiable} at
  $s$. In this case $s$ is labeled as a ``singular'' switch point or a
  ``\emph{dynamic singularity}'' (previous works labeled such switch
  points as ``zero-inertia points''~\cite{SY89tra}; however we shall
  see below that not all zero-inertia points are singular);
\item the $\MVC$ is \emph{continuous} and \emph{differentiable} at $s$
  and the \emph{tangent vector} to the $\MVC$ at $(s,\MVC(s))$ is
  collinear with the vector $(\MVC(s),\alpha(s,\MVC(s)))$ [or, which
  is the same since we are on the $\MVC$, collinear with the vector
  $(\MVC(s),\beta(s,\MVC(s)))$]. In this case $s$ is a labeled as a
  ``tangent'' switch point.
\end{itemize}

Finding discontinuous and tangent switch points does not involve
particular difficulties since it suffices to construct the $\MVC$ and
examine whether it is discontinuous or whether the tangent to the
$\MVC$ is collinear with $\alpha$ for all discretized points $s$ along
the path. Regarding the undifferentiable switch points, one approach
could consist in checking whether the $\MVC$ is continuous but
undifferentiable at $s$. However, this approach is seldom used in
practice since it is comparatively more prone to discretization
errors. Instead, it was proposed
(cf.~\cite{PJ87jra,SY89tra,SL92jdsmc,ChoX05book,KS12rss}) to equate
undifferentiable points with \emph{zero-inertia points}, i.e. the
points $s$ where $a_k(s)=0$ for one of the constraints $k$, and to
consequently search for zero-inertia points.

This method is however not completely correct: we shall see in
Section~\ref{sec:charact} that \emph{not all zero-inertia points are
  undifferentiable}.
%
%
Thus, only zero-inertia points that are undifferentiable properly
constitute \emph{singular} switch points or \emph{dynamic
  singularities}. Characterizing and addressing such switch points are
of crucial importance since they occur in a large proportion of
real-world TOPP instances -- they are in particular much more frequent
than discontinuous and tangent switch points together. Furthermore,
since the $\alpha$ and $\beta$ fields are \emph{divergent} near these
switch points (see Fig.~\ref{fig:flows}), they are the cause of most
failures in existing implementations of TOPP.
 
\subsection{Characterizing dynamic singularities}
\label{sec:charact}

Consider a zero-inertia point $s^*$, and assume that it is triggered
by the $k$-th constraint, i.e. $a_k(s^*)=0$. Without loss of
generality, we make the assumption that $a_k(s)<0$ in a neighborhood
to the left of $s^*$ and $a_k(s)>0$ in a neighborhood to the right of
$s^*$ (the case when $a_k$ switches from positive to negative can be
treated similarly by changing signs at appropriate places).

\begin{figure}[htp]
    \begin{minipage}[hc]{4.5cm}
    \includegraphics[width=4.5cm]{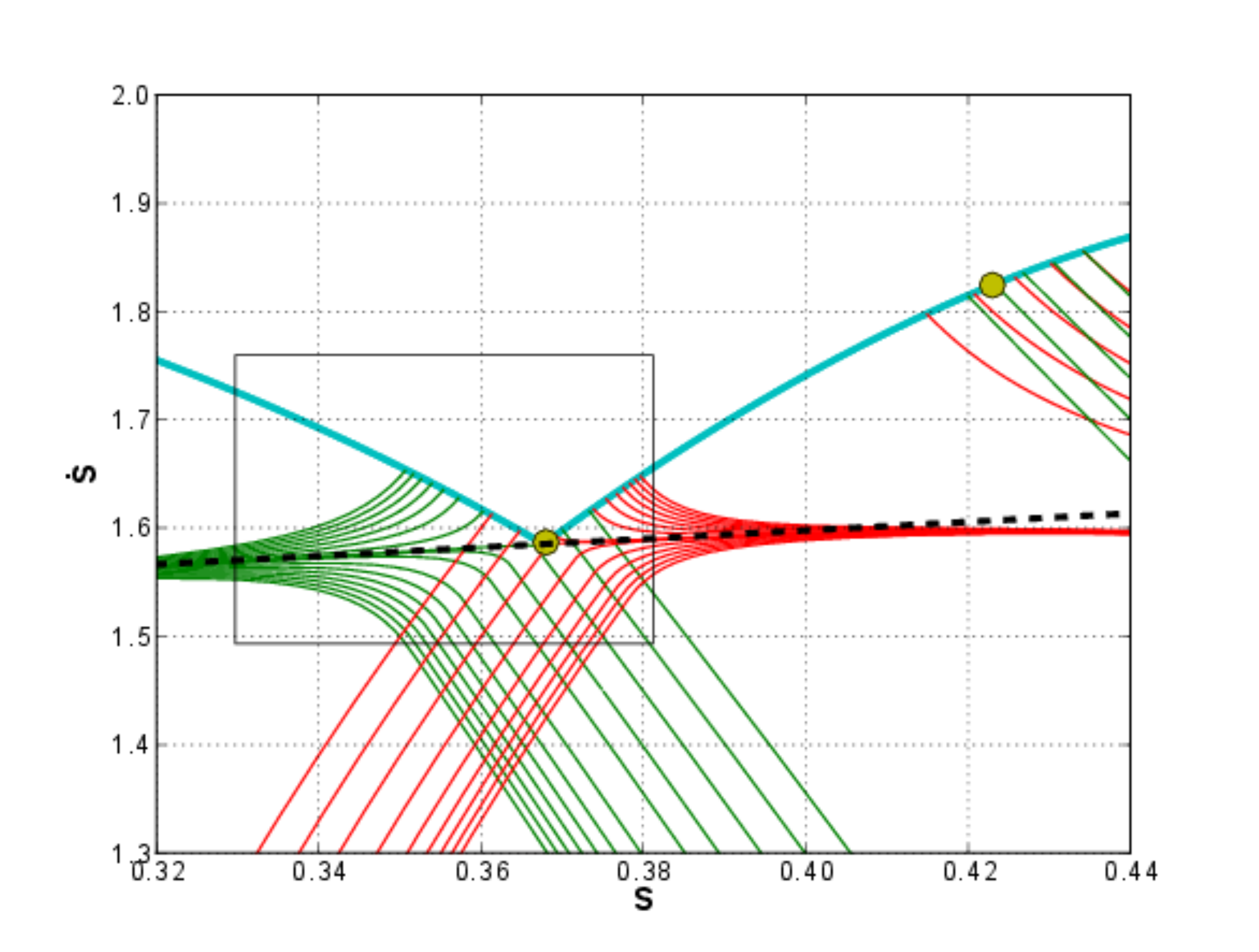}
    \end{minipage}
    \begin{minipage}[hc]{4cm}
      \includegraphics[width=4cm]{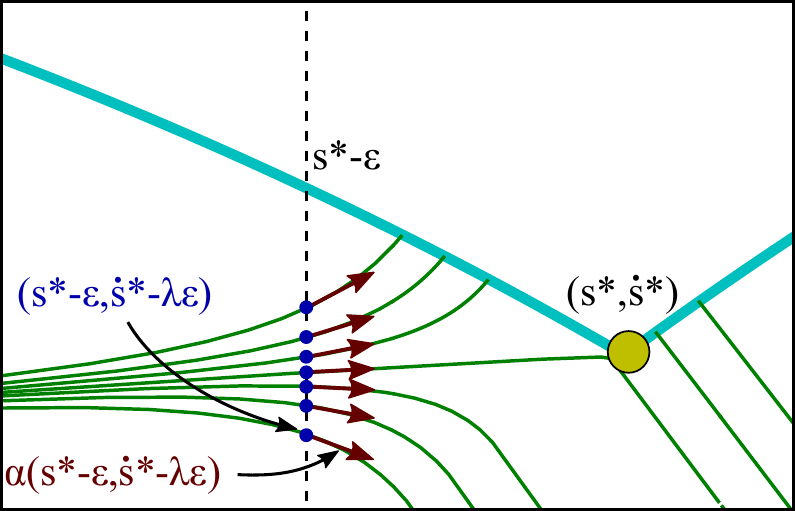}
    \end{minipage}
    \caption{\textbf{A}: $\alpha$- and $\beta$-profiles (in green and
      red respectively) near zero-inertia points (yellow points). The
      left zero-inertia point is a singular switch point, while the
      right zero-inertia point is not singular. Note that in agreement
      with the definitions, at any point in the plane the slope of the
      red profile is higher than the slope of the green profile,
      except on the MVC where the two slopes are equal. The dotted
      line is the line that goes through the switch point and has
      slope $\lambda$ computed by equation~(\ref{eq:lambda})
      (cf. Section~\ref{sec:addr}). \textbf{B}: close-up view (zoomed
      in the black box of \textbf{A}) centered around the singular
      switch point.}
  \label{fig:flows}
\end{figure}

We prove in the Appendix that, if the path is traversable, then
$c_k(s^*)<0$. We next distinguish two cases (see
Fig.~\ref{fig:flows}A).

\textbf{Case 1: $b_k(s^*)>0$}. Define
$\dot{s}^*=\sqrt{\frac{-c_k(s^*)}{b_k(s^*)}}$. Next, let
$\dot{s}^\dag{}$ be the value of the MVC, had we removed constraint
$k$. We prove in the Appendix that
\begin{itemize}
\item If $\dot{s}^\dag{}<\dot s^*$, then constraint $k$ does
  \emph{not} trigger a dynamic singularity at $s^*$;
\item If $\dot{s}^\dag{}>\dot s^*$, then the MVC is indeed
  undifferentiable at $s^*$, and $s^*$ constitutes a dynamic
  singularity.
\end{itemize}

\textbf{Case 2: $b_k(s^*)<0$}. We prove that, in this case, constraint
$k$ does \emph{not} trigger a dynamic singularity at $s^*$.

\subsection{Addressing dynamic singularities}
\label{sec:addr}

\subsubsection{Previous treatments}

The next difficulty consists in the selection of the optimal
acceleration to initiate the backward and forward integrations from a
dynamic singularity $s^*$: indeed the fields $\alpha$ and $\beta$ are
not \emph{naturally} defined at these points because of a division by
$a_k(s^*)=0$. In~\cite{PJ87jra}, no indication was given regarding
this matter. In~\cite{SY89tra}, it was stated that ``\emph{[this]
  acceleration is not uniquely determined}'' and suggested to choose
\emph{any} acceleration to initiate the
integrations. In~\cite{SL92jdsmc} (and also
in~\cite{ChoX05book} which reproduced the reasoning
of~\cite{SL92jdsmc}), the authors suggested to select the
following acceleration to initiate the backward integration (and a
similar expression for the forward integration):
\begin{equation}
  \label{eq:shiller}
  \min(\alpha^-,\alpha^+,\alpha_\MVC),\ \textrm{where}
\end{equation}
\begin{eqnarray}
  \label{eq:alpha}
  \alpha^-=\lim_{s\uparrow s^*} \alpha(s,\MVC(s^*)),\
  \alpha^+=\lim_{s\downarrow s^*} \alpha(s,\MVC(s^*)),
\end{eqnarray}
and $\alpha_\MVC$ is computed from the \emph{slope} of the $\MVC$
on the left of $s^*$.

However, observing the $\alpha$-profiles near the dynamic singularity
of Fig.~\ref{fig:flows}A, it appears that the definition of $\alpha^-$
in equation~(\ref{eq:alpha}) is arbitrary. Indeed, depending on the
\emph{direction} from which one moves towards $(s^*,\MVC(s^*))$ in the
$(s,\dot s)$ plane, the limit of $\alpha$ is different: for instance,
in Fig.~\ref{fig:flows}A, if one moves from the top left, the limit,
if it exists, would be positive, and it would be negative if one moves
from the bottom left. In this context, the choice of
equation~(\ref{eq:alpha}) consisting in moving towards
$(s^*,\MVC(s^*))$ \emph{horizontally} is no more justified than any
other choice. More generally, it is \emph{impossible} to extend
$\alpha$ by continuity towards $(s^*,\MVC(s^*))$ from the left because
the $\alpha$-profiles \emph{diverge} when approaching
$(s^*,\MVC(s^*))$ from the left.

In practice, because of this flow divergence, choosing a slightly
incorrect value for $\alpha$ and $\beta$ at $s^*$ may result in strong
oscillations (see Fig.~\ref{fig:compare}), which in turn can make the
algorithm incorrectly terminate (because the the velocity profile
would cross the MVC or the line $\dot s=0$). In fact, this is probably
one of the main reasons of failure in existing implementations.

\subsubsection{Proposed new treatment}

Fig.~\ref{fig:flows}B shows in more detail the $\alpha$-profiles near
a singular switch point $s^*$. 

We first show in the Appendix that, if the singularity is triggered by
constraint $k$, then $\alpha=\alpha_k$ on the left of $s^*$ and
$\beta=\beta_k$ on the right of $s^*$. Consider next the intersections
of the vertical line $s=s^*-\epsilon$, where $\epsilon$ is an
arbitrary small positive number, with the $\alpha$-profiles. An
$\alpha$-profile can reach $(s^*,\dot s^*)$ only if its \emph{tangent
  vector} at the intersection \emph{points towards} $(s^*,\dot
s^*)$. This can be achieved if there exists a real number $\lambda$
such that
\[
\frac{\alpha_k(s^*-\epsilon,\dot s^*-\lambda\epsilon)}{\dot s^*-\lambda\epsilon}=\lambda.
\]
Replacing $\alpha_k$ by its expression yields the condition
\[
\frac{-b_k(s^*-\epsilon)[\dot
  s^*-\lambda\epsilon]^2-c_k(s^*-\epsilon)}{a_k(s^*-\epsilon)[\dot
  s^*-\lambda\epsilon]}=\lambda,\ \textrm{i.e.
}
\]
\[
-b_k(s^*-\epsilon)[\dot
  s^*-\lambda\epsilon]^2-c_k(s^*-\epsilon)=\lambda a_k(s^*-\epsilon)[\dot s^*-\lambda\epsilon].
  \]
Computing the Taylor expansion of the above equation at order 1 in
$\epsilon$ and recalling that $-b_k(s^*)\dot
s^{*2}-c_k(s^*)=0$ and $a_k(s^*)=0$, one obtains the condition
\[
2\lambda b_k(s^*)\dot s^* + b'_k(s^*)\dot{s}^{*2}+
c'_k(s^*)=-\lambda  a'_k(s^*) \dot s^*.
\]
Solving for $\lambda$, one finally obtains
\begin{equation}
  \label{eq:lambda}
  \lambda=-\frac{b'_k(s^*)\dot{s}^{*2}+c'_k(s^*)}{[2
    b_k(s^*) +a'_k(s^*)]\dot s^*}.
\end{equation}

Following the same reasoning on the right of $s^*$, one has to solve
\[
\frac{\beta_k(s^*+\epsilon,\dot s^*+\lambda\epsilon)}{\dot s^*+\lambda\epsilon}=-\lambda,
\]
which leads to the same value as in equation~(\ref{eq:lambda}). Thus
the optimal backward and forward acceleration at $(s^*,\dot s^*)$ is
given by equation~(\ref{eq:lambda}). One can observe in
Fig.~\ref{fig:flows}A that the black dotted line, whose slope is given
by $\lambda$, indeed constitutes the ``neutral'' line at $(s^*,\dot
s^*)$.

Based on the previous development, we propose the following algorithm
when encountering a zero-inertia point~$s^*$, with $a_k(s^*)<0$ on the
left of $s^*$ and $a_k(s^*)>0$ on the right of $s^*$:

\noindent \textbf{$\blacktriangle$}\,If $b_k(s^*)<0$, then constraint
$k$ does not trigger a singularity;

\noindent \textbf{$\blacktriangle$}\,If $b_k(s^*)>0$, then compute $\dot{s}^*$ by
  equation~(\ref{eq:sdotstar}) and $\dot{s}^\dag$ by removing
  constraint $k$ and evaluating again the MVC at $s^*$.
  \begin{itemize}
  \item If $\dot{s}^*>\dot{s}^\dag$, then constraint $k$ does not
    trigger a singularity;
  \item If $\dot{s}^*<\dot{s}^\dag$, then $s^*$ is a dynamic
    singularity. Next, compute $\lambda$ by equation~(\ref{eq:lambda})
    and
    \begin{itemize}
    \item integrate the constant field $(\dot s^*,\lambda\dot s^*)$
      \emph{backward} for a small number of time steps. Then continue by
      following $\alpha$, as in the original algorithm;
    \item integrate the constant field $(\dot s^*,\lambda\dot s^*)$
      \emph{forward} for a small number of time steps. Then continue by
      following $\beta$.
    \end{itemize}
  \end{itemize}

Note that after moving a small number of steps away from $s^*$, the
fields $\alpha$ and~$\beta$ become smooth, so that there is no problem
in the integration.

\subsubsection{About Kunz and Stilman's conjecture}

Kunz and Stilman~\cite{KS12rss} were first to remark -- in the
particular case of TOPP with velocity and acceleration bounds and
paths made of straight segments and circular arcs -- that the
algorithm proposed in~\cite{SL92jdsmc} could not satisfactorily
address dynamic singularities. They conjectured instead that the
correct acceleration at the singularity is 0.

From equation (14) of~\cite{KS12rss}, the correspondences between the
parameters of~\cite{KS12rss} and those of the present article are
given in Table~\ref{tab:corresp}.

\begin{table}[th]
  \caption{Parameters correspondences}    
  \centering
    \begin{tabular}{|ccc|}
      \hline
      This article&&Kunz and Stilman~\cite{KS12rss}\\
      \hline      
      $a_k(s)$&$\leftrightarrow$&$f_k'(s)$\\
      $b_k(s)$&$\leftrightarrow$&$f_k''(s)$\\
      $c_k(s)$&$\leftrightarrow$&$-\ddot q_k^{\max}$ or $\ddot q_k^{\max}$\\
      \hline
     \end{tabular}
 \label{tab:corresp}  
\end{table}

Remark  that the zero-inertia points in~\cite{KS12rss} are all
located in the circular portions. In such portions, the coefficients
$a_k$ and $b_k$ have the following form (using our notations):
{\begin{eqnarray}
  \label{eq:toto}
  a_k(s)&=&-\frac{C_1}{r} \sin\left(\frac{s}{r}\right)+\frac{C_2}{r} \cos\left(\frac{s}{r}\right),\nonumber\\
  b_k(s)&=&\frac{C_1}{r^2} \cos\left(\frac{s}{r}\right)-\frac{C_2}{r^2} \sin\left(\frac{s}{r}\right),\nonumber
\end{eqnarray}}
where $r$, $C_1$ and $C_2$ are three constants independent of $s$ in a
neighborhood around $s^*$.
Differentiating $b_k$ next yields
{
\[
b'_k(s)=-\frac{C_1}{r^3} \sin\left(\frac{s}{r}\right)+\frac{C_2}{r^3} \cos\left(\frac{s}{r}\right)=\frac{1}{r^2} a_k(s).
\]}
One thus has $b'_k(s^*)=1/r^2a_k(s^*)=0$ at a zero-inertia point. If
this zero-inertia point is actually a dynamic singularity then, from
equation~(\ref{eq:lambda}), one obtains that $\lambda=0$, which proves
Kunz and Stilman's conjecture.

\section{Implementation and evaluation}
\label{sec:implementation}

\subsection{Open-source implementation}

We provide an implementation of TOPP in C++ that integrates the
developments of Section~\ref{sec:robust}.  We also provide an
interface in Python for an easy and interactive use. Currently, our
implementation supports pure velocity and acceleration bounds, torque
bounds, ZMP constraints, multi-contact friction constraints, and any
combination thereof. The dynamics computations are handled by
OpenRAVE~\cite{Dia10these}. Thanks to the general formulation of
Section~\ref{sec:general}, new system dynamics and constraints can be
easily added. The implementation and test cases are open-source and
available at \url{https://github.com/quangounet/TOPP/releases}.

\begin{figure}[htp]
    \centering
    \hspace{-0.1cm}
    \includegraphics[width=2cm]{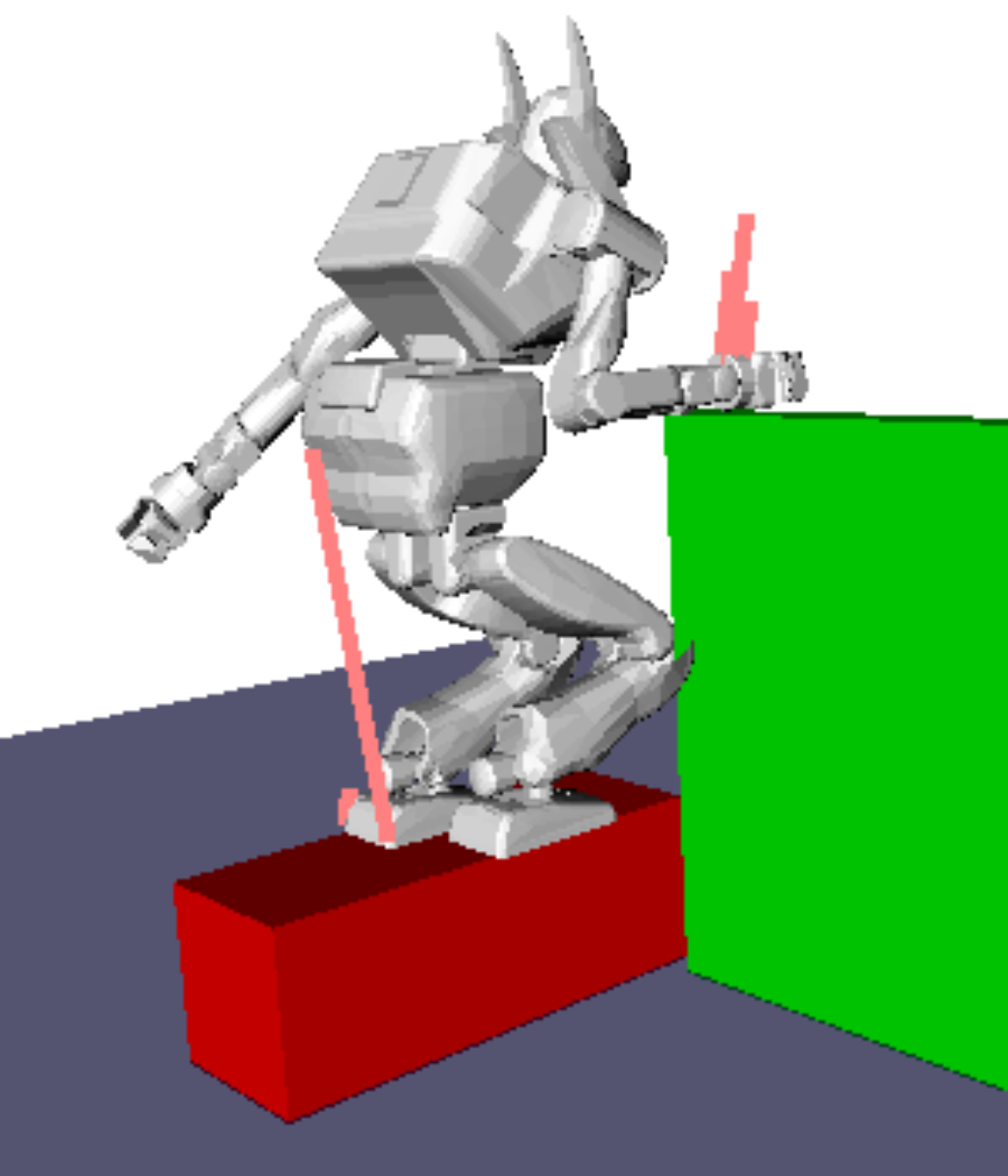}
    \hspace{-0.1cm}
    \includegraphics[width=2cm]{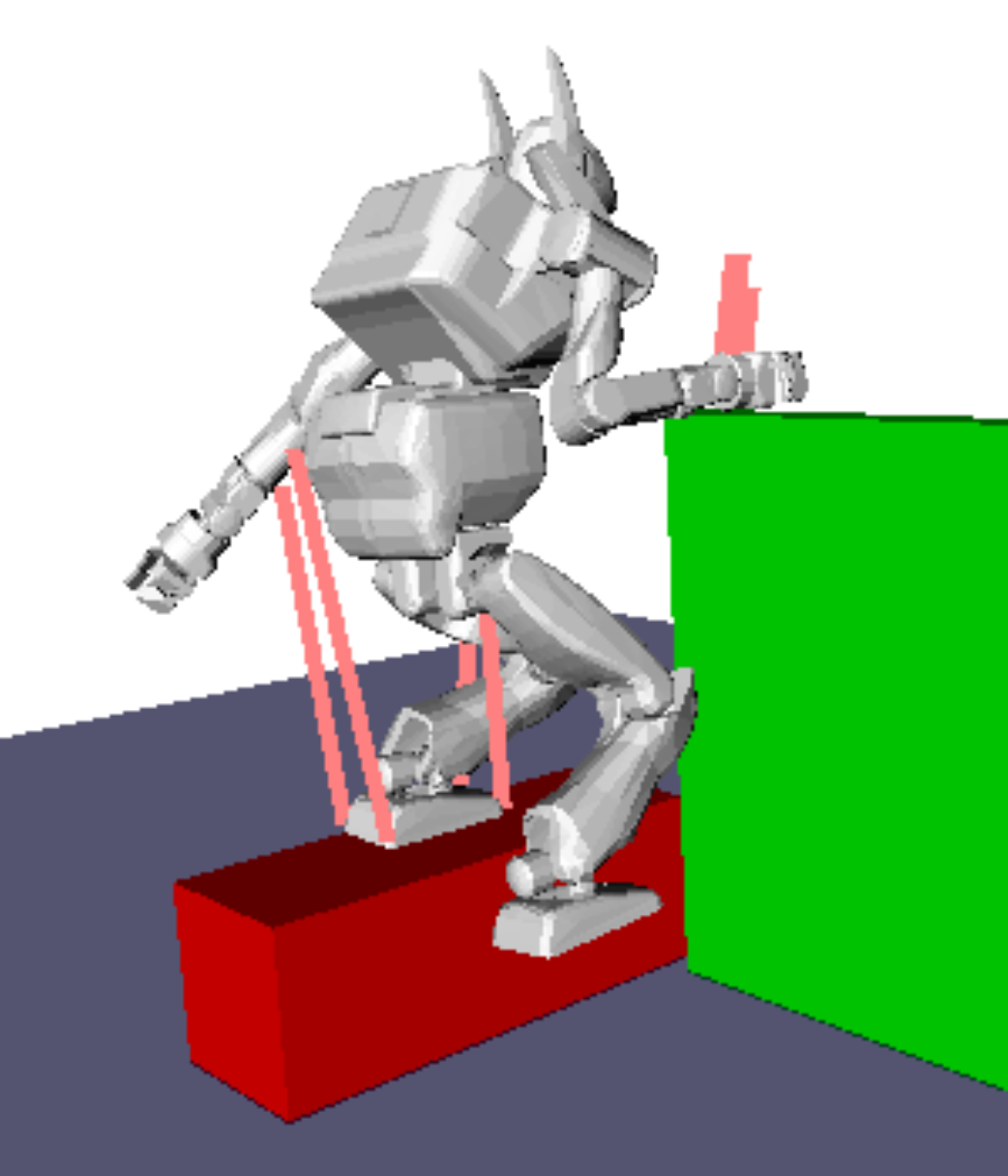}
    \hspace{-0.1cm}
    \includegraphics[width=2cm]{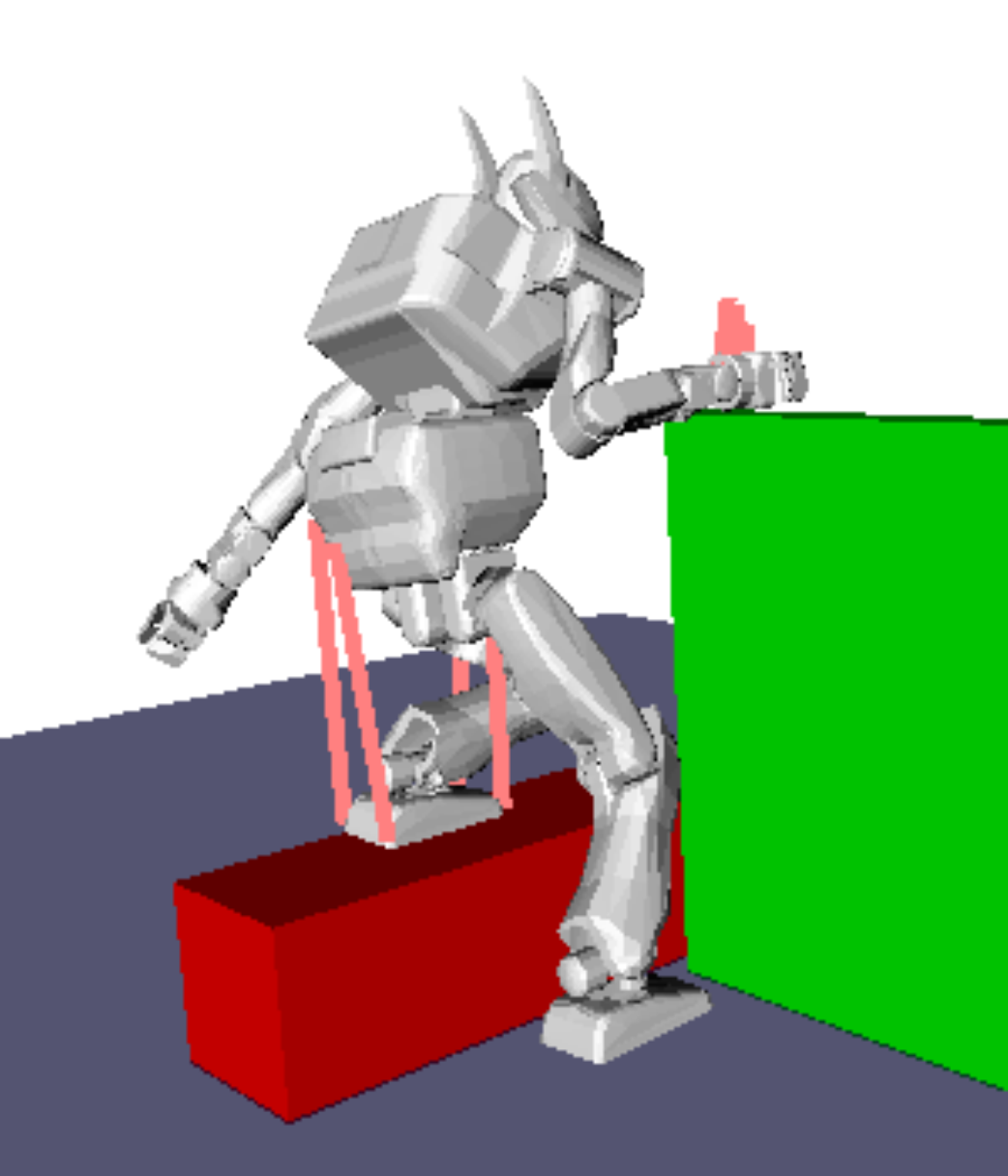}
    \hspace{-0.1cm}
    \includegraphics[width=2cm]{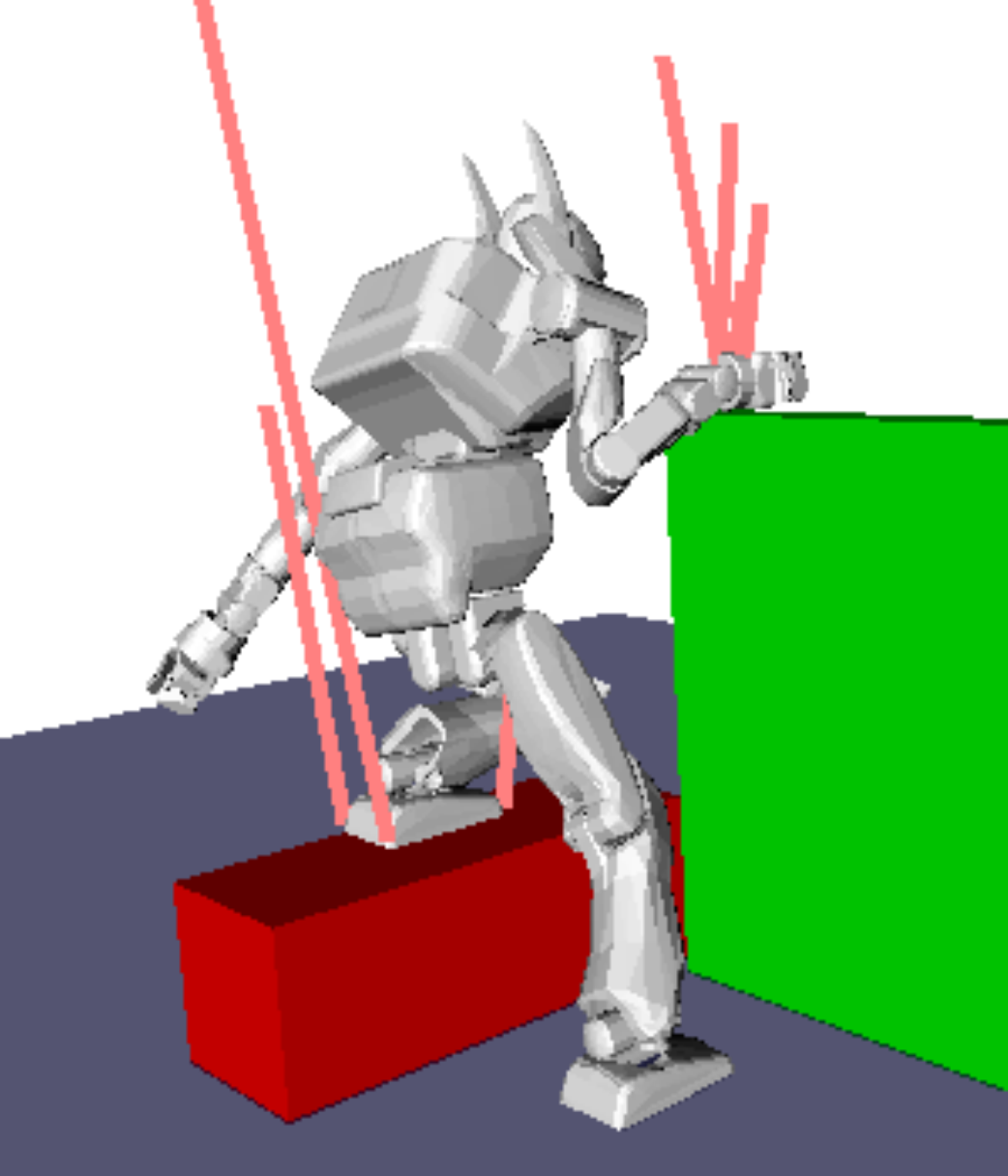}\\
    \hspace{-0.3cm}
    \includegraphics[width=4.4cm]{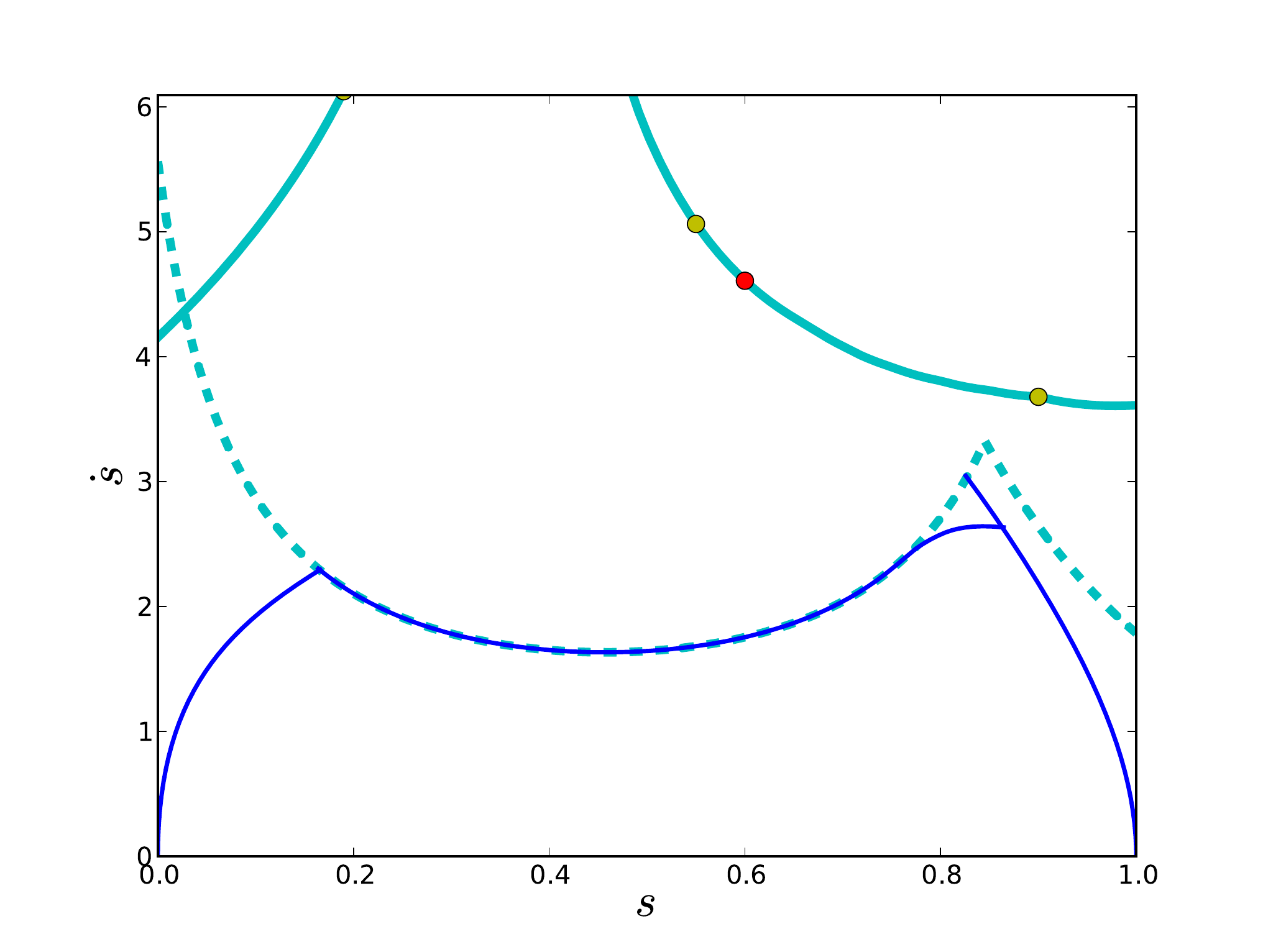}
    \hspace{-0.3cm}
    \includegraphics[width=4.4cm]{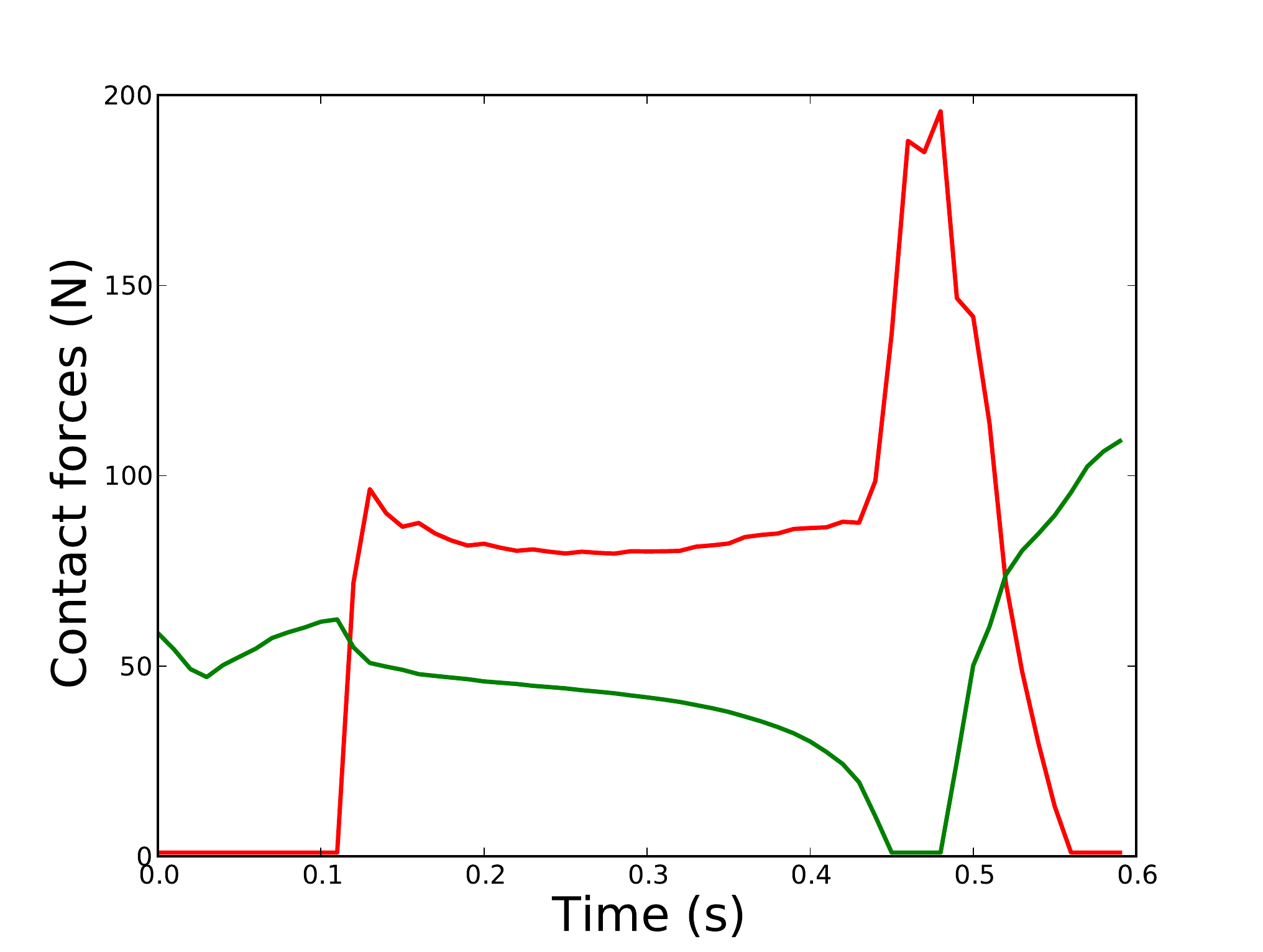}
    \caption{TOPP with velocity, torque, and friction constraints in a
      multi-contact task with the HRP2 robot. \textbf{Top}: snapshots
      of the time-parameterized trajectory taken at equal time
      intervals. The pink lines represent the contact
      forces. \textbf{Left}: $(s,\dot s)$ space. Same legends as in
      Fig.~\ref{fig:flows}. The solid and dotted bold cyan lines are
      velocity limits that are imposed respectively by the torque and
      friction constraints [the MVC computed from
      equation~(\ref{eq:gen})] and by the ``direct'' velocity
      constraints [computed from equation~(\ref{eq:velo})]. The
      superimposed dotted blue line represents the final $(s,\dot s)$
      profile, which for some parts followed the $\alpha$- and
      $\beta$-profiles and for some other parts ``slid'' along the
      ``direct'' velocity limit. \textbf{Right}: normal components of
      the reaction forces at the front left corner of the left foot
      (red) and the front left corner of the right hand (green). The
      normal components were constrained to be $\geq 1$\,N. Note how
      this constraint was saturated at the foot and hand contact
      points at different moments in time.}

    \label{fig:humanoid}
\end{figure}

Fig.~\ref{fig:humanoid} illustrates the utilization of TOPP in a
multi-contact task where a humanoid robot (HRP2, 36 dofs including the
6 coordinates of the free-flyer) steps down from a 30\,cm high
podium. Velocity and torque bounds were considered for each joint, as
well as friction cone constraints on the left foot and the right hand
(131 inequality constraints in total). The original trajectory had
duration 1\,s and the grid size was $N=100$. The time parameterization
part (excluding the dynamics computations and the constraints
projection step~\cite{Hau14icra}) took 0.005\,s on our computer (Intel
Core i5 3.2GHz, 3.8GB memory, GNU/Linux).

\subsection{Comparison with previous treatments of dynamic
  singularities}
\label{sec:compshiller}

\begin{figure}[htp]
    \centering
    \hspace{-0.5cm}
    \includegraphics[width=2cm]{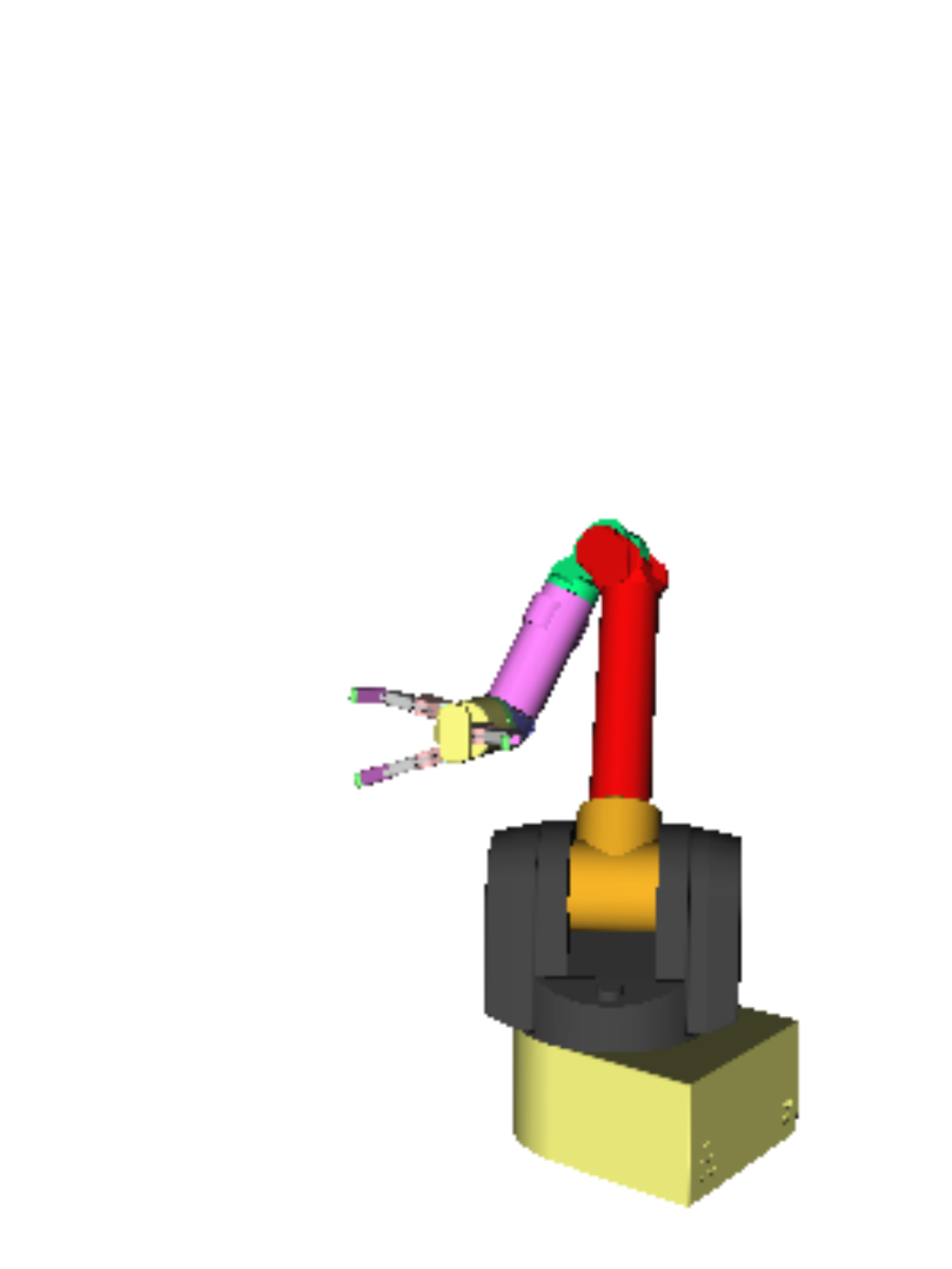}
    \hspace{-0.5cm}
    \includegraphics[width=2cm]{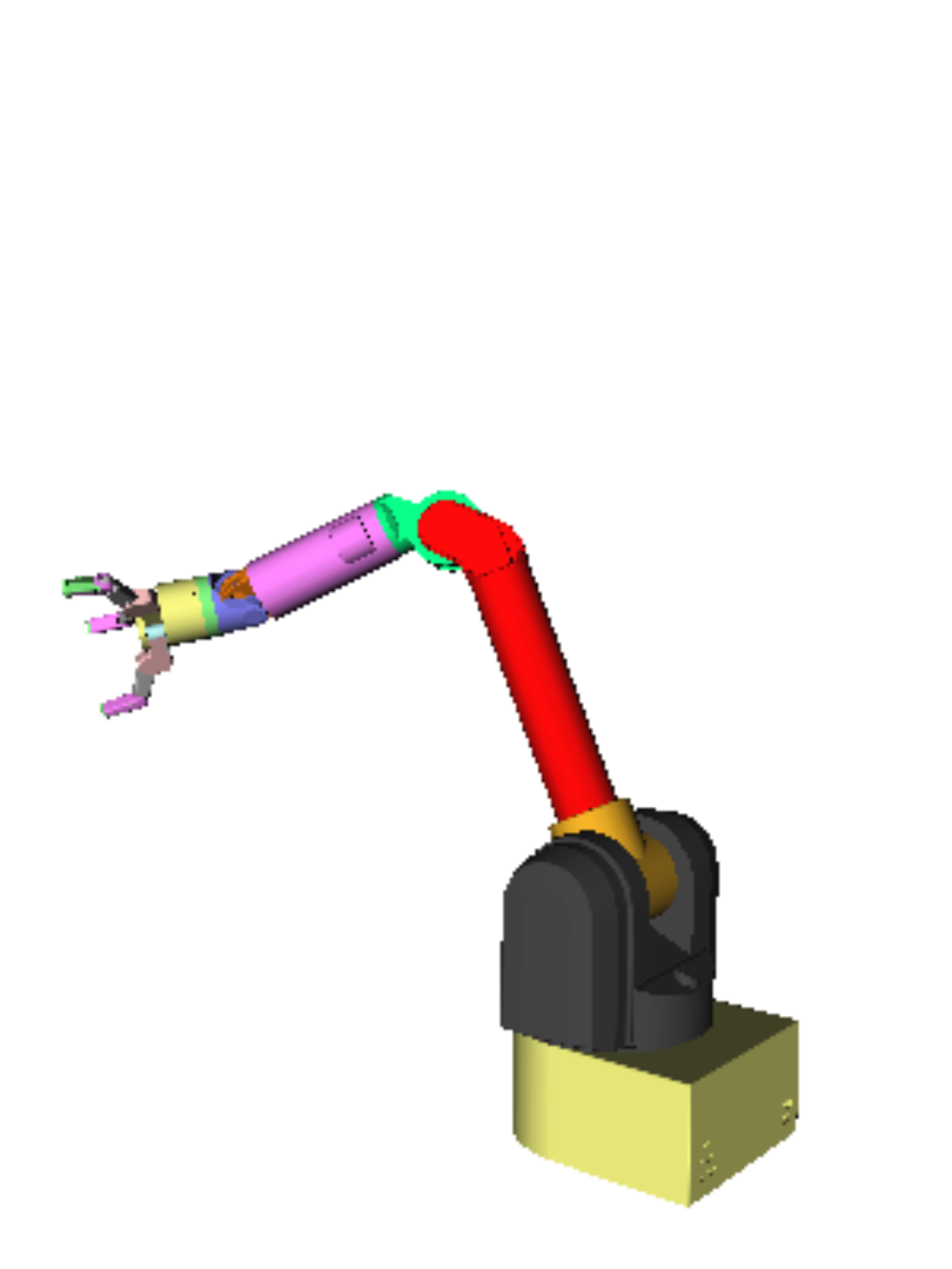}
    \hspace{-0.5cm}
    \includegraphics[width=2cm]{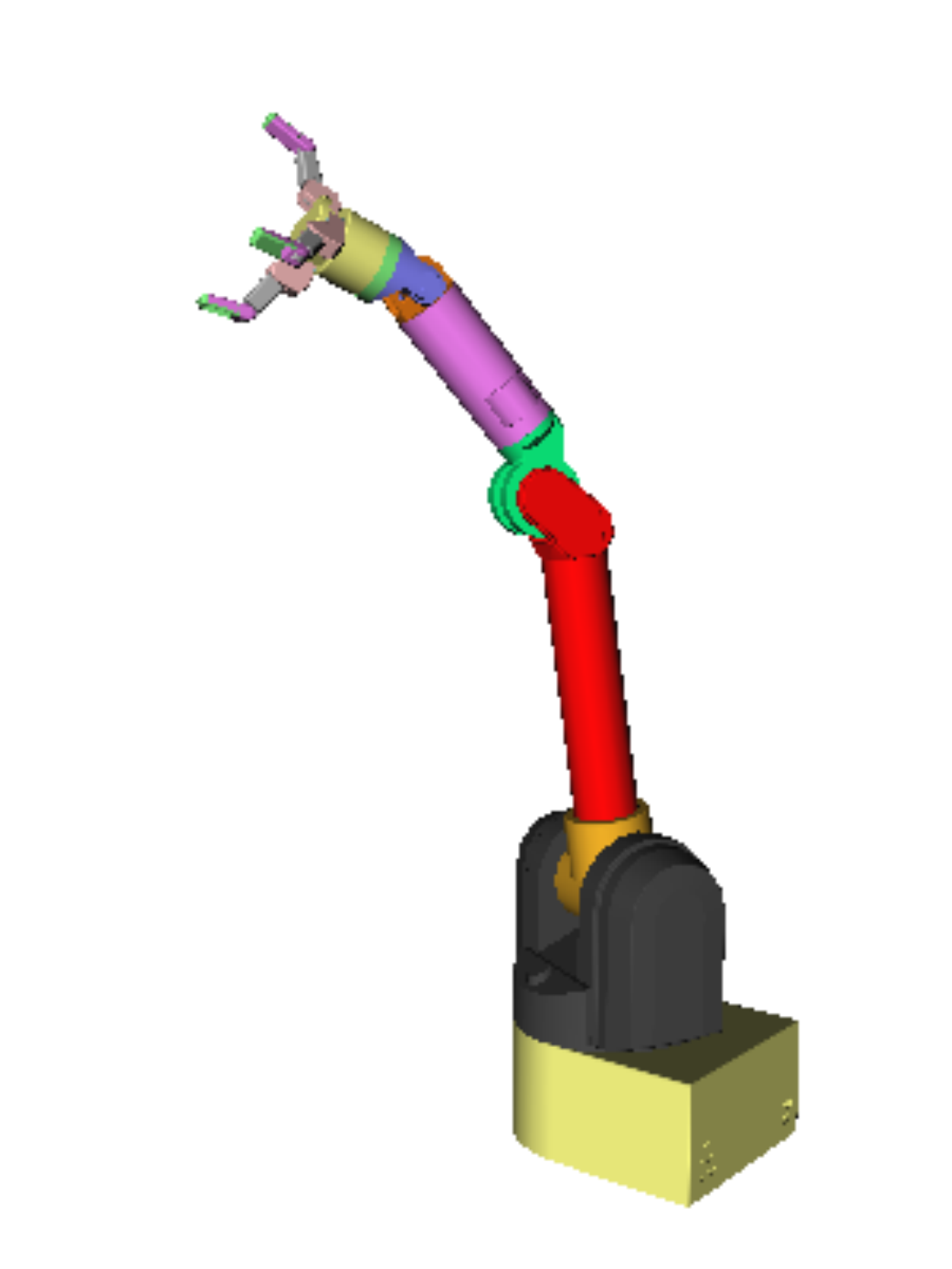}
    \hspace{-0.5cm}
    \includegraphics[width=2cm]{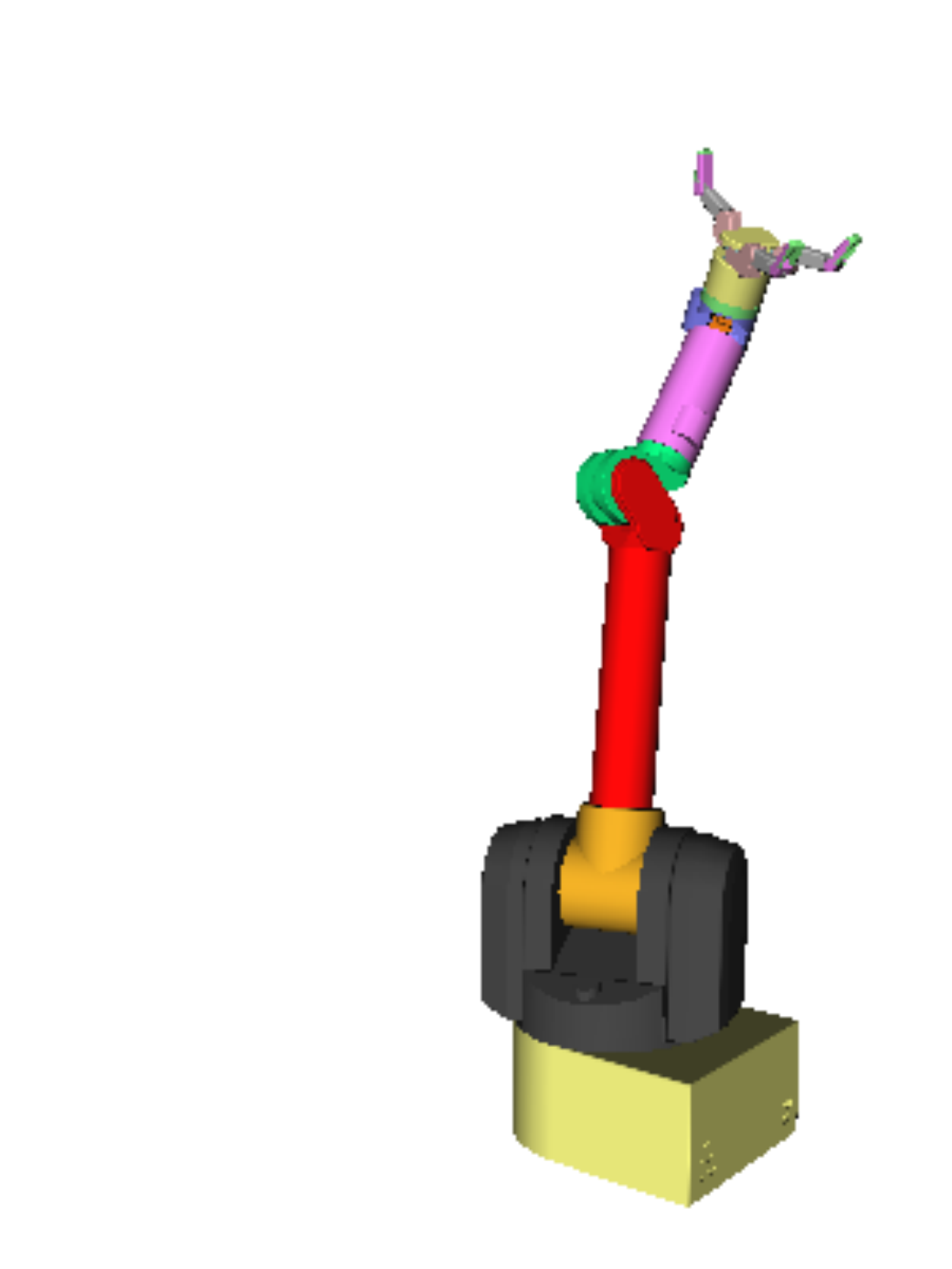}
    \hspace{-0.5cm}
    \includegraphics[width=2cm]{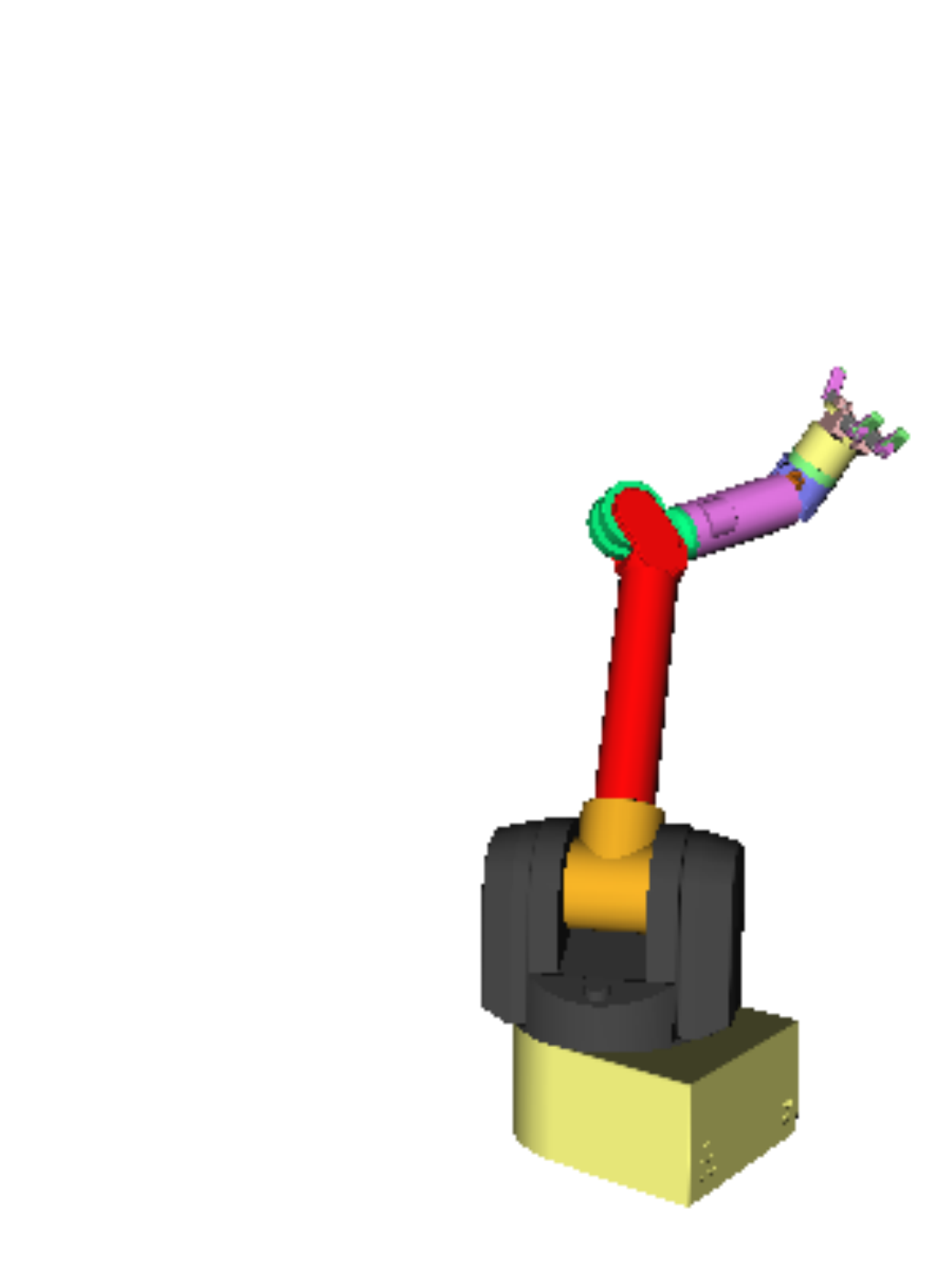}\\
    \hspace{-0.3cm}
    \includegraphics[width=4.4cm]{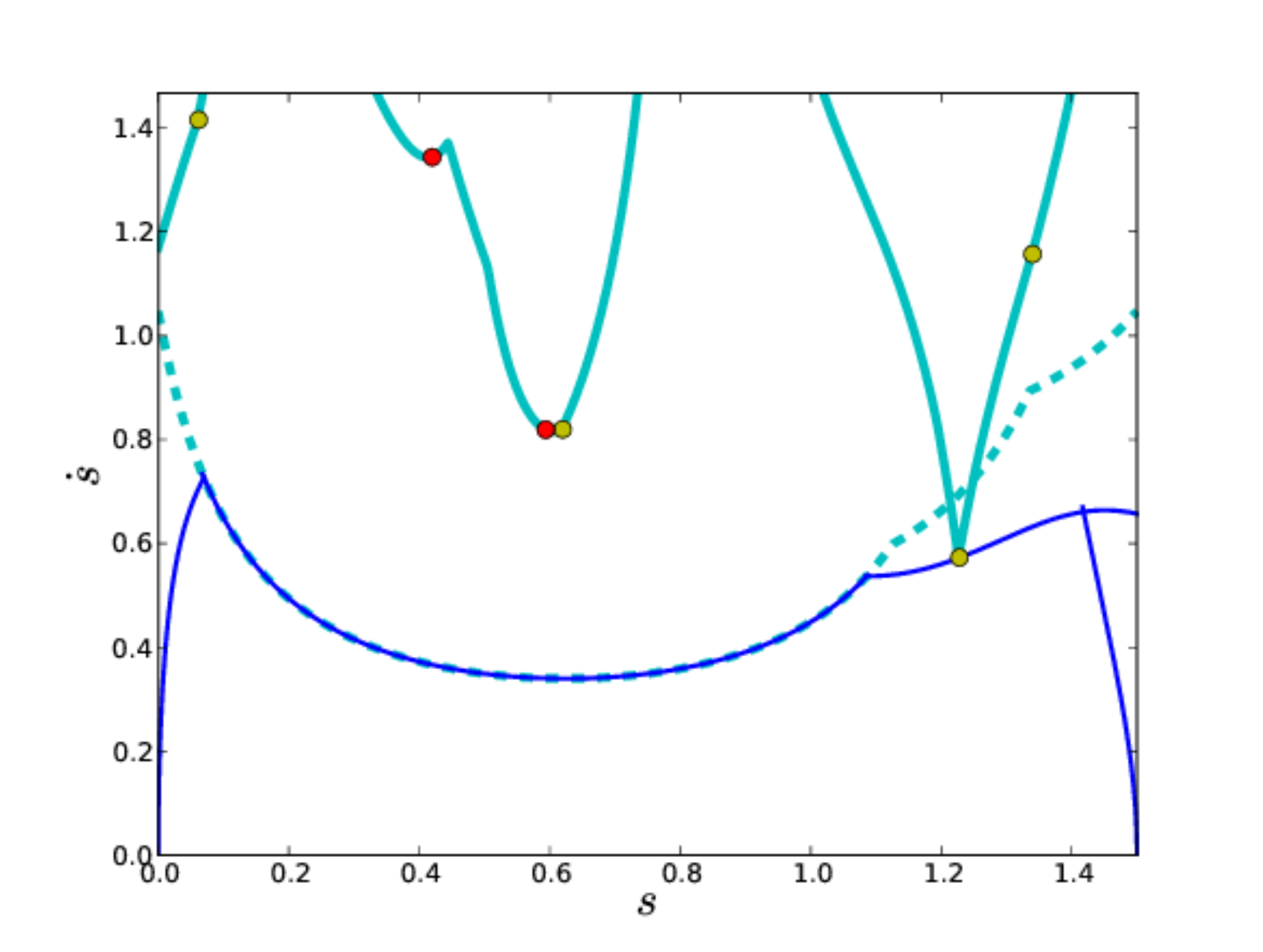}
    \hspace{-0.3cm}
    \includegraphics[width=4.4cm]{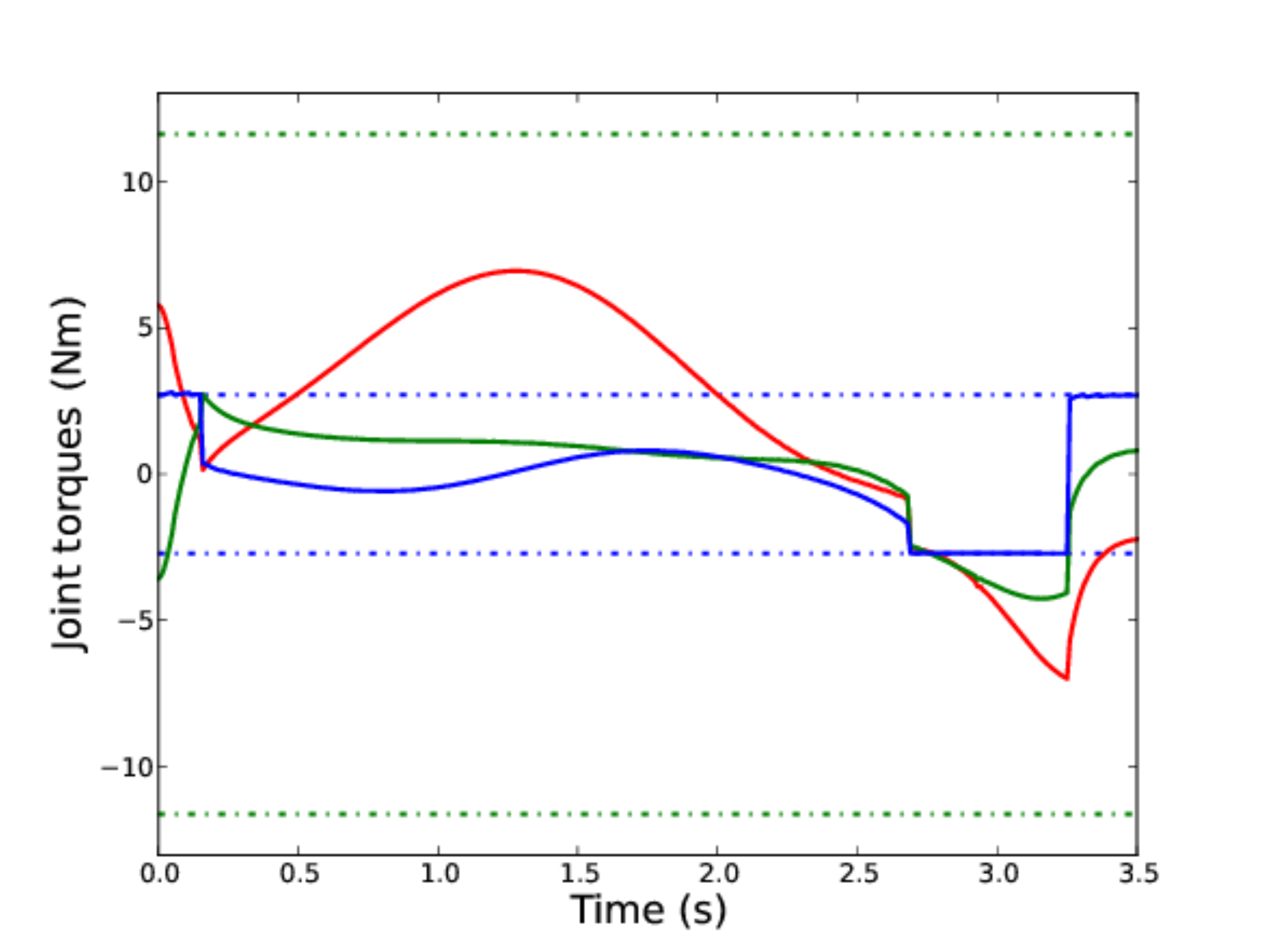}
    \caption{TOPP with velocity and torque constraints for the 7-dof
      Barrett WAM. \textbf{Top}: snapshots of the time-parameterized
      trajectory taken at equal time intervals. \textbf{Left}:
      $(s,\dot s)$ space. Same legends as in
      Fig.~\ref{fig:humanoid}. \textbf{Right}: torque profiles for
      shoulder roll (solid red), wrist yaw (solid green) and wrist
      roll (solid blue). The dotted lines represent the torque bounds
      in corresponding colors.}
    \label{fig:algo}
\end{figure}

For this, we tested the algorithm on a model of the 7-dof Barrett
WAM. The velocity and torque bounds are those given by the
constructor. Fig.~\ref{fig:algo} shows a smooth behavior around the
dynamic singularity, for both the $(s,\dot s)$ profile and the torque
profiles.

\begin{figure}[htp]
    \centering
    \includegraphics[width=4.4cm]{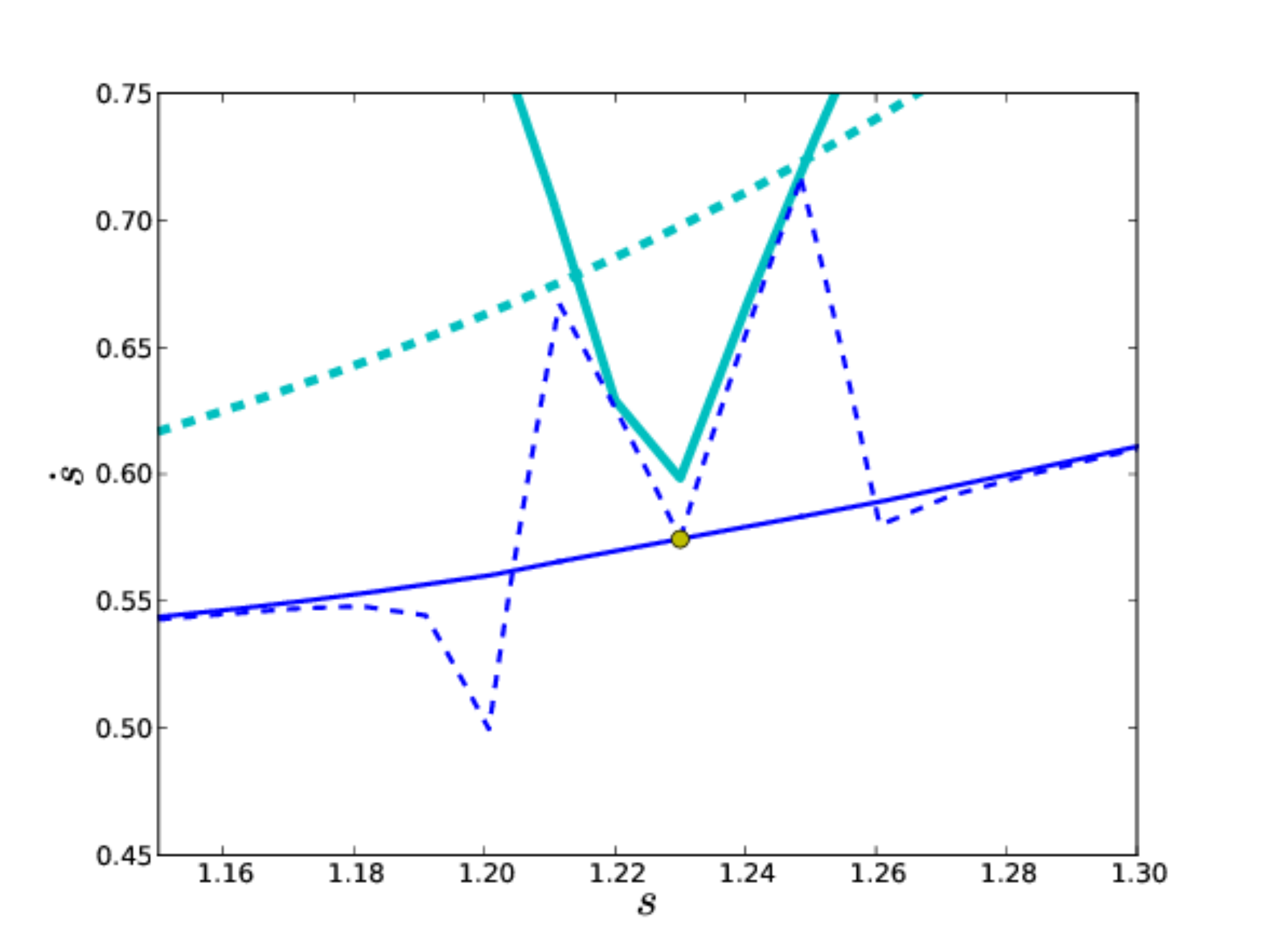}
    \hspace{-0.5cm}
    \includegraphics[width=4.4cm]{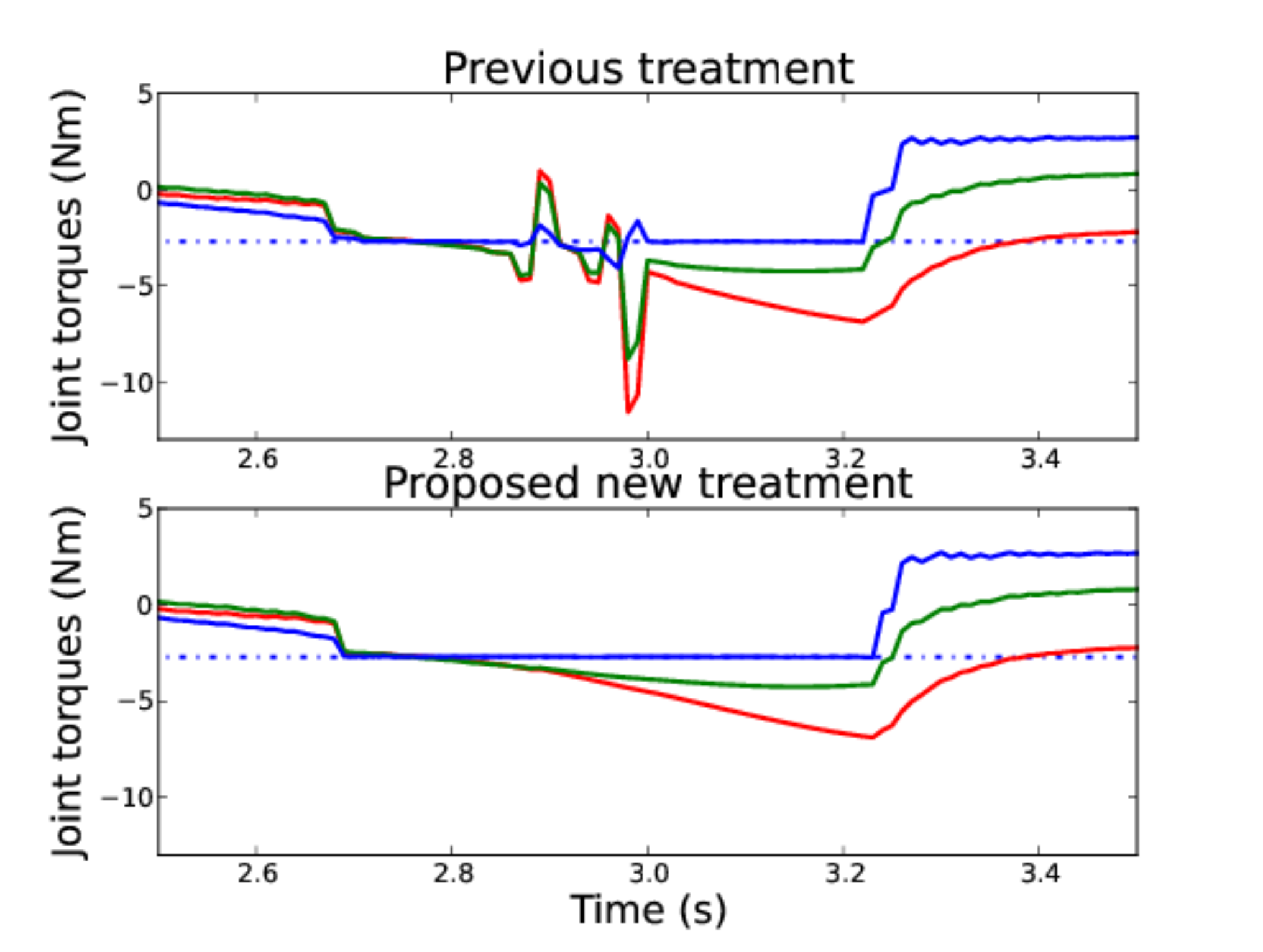}\\
    \caption{Close-up views of the $(s,\dot s)$ profiles and the
      torque profiles near a dynamic singularity. The computations
      were done at a time step of 0.01\,s. \textbf{Left}: $(s,\dot s)$
      profiles. Dotted lines: profile computed using the method
      of~\cite{SL92jdsmc}. Solid lines: profile computed using our
      proposed new method. \textbf{Right, top}: torque profiles
      corresponding to the method of~\cite{SL92jdsmc}. \textbf{Right,
        bottom}: torque profiles corresponding to our method. Note
      that our method allows suppressing the jitters even at a coarse
      time step.}
    \label{fig:compare}
\end{figure}

Next, to demonstrate more clearly the improvements permitted by our
algorithm, we compared the results given by our algorithm and that
given by the algorithm of~\cite{SL92jdsmc}, which incorrectly proposes
to ``slide'' along the MVC near dynamic singularities.
Fig.~\ref{fig:compare} shows that using the correct acceleration
values significantly decreases the jitters around the dynamic
singularity, even at a coarse discretization time step.

Finally, we tested the overall robustness of the implementation in two
settings: the 7-dof WAM with velocity and torque bounds as above, and
a 7-dof kinematic system with velocity and acceleration bounds (set to
respectively 4\,rad$\cdot$s$^{-1}$ and 20\,rad$\cdot$s$^{-2}$, which
are typical values for industrial manipulators).  In both settings, we
tested 1000 random trajectories of duration 1\,s and grid size
$N=200$. The initial trajectories were Bezier curves that interpolate
between two random points in $[-\pi,\pi]^7$. Table~\ref{tab:robust}
reports the number of failures -- which occur when the profiles do not
cover the whole segment $[0,s_\fin]$~\cite{KS12rss} -- as well as the
number of dynamic singularities encountered in the 1000 instances.

\begin{table}[th]
  \caption{Testing the implementation on 1000 trajectories}    
  \centering
    \begin{tabular}{|l|c|c|}
      \hline
      System&$\sharp$ failures& $\sharp$ singularities\\
      \hline      
      WAM torque constraints&0&307\\
      Kinematic constraints&1&539\\
      \hline
     \end{tabular}
 \label{tab:robust}  
\end{table}

\subsection{Comparison with the convex optimization approach}
\label{sec:compconvex}

We first conducted an ``informal'' comparison of our implementation
with the MATLAB-based implementations of Verscheure et
al~\cite{VerX09tac} and of Debrouwere et al~\cite{DebX13tr} on the
problem of torque bounds. Note that, for the convex optimization part
itself, these implementations used a library, YALMIP, which in turn
calls an external solver (SeDuMi in~\cite{VerX09tac}), written in C,
C++ or Fortran and precompiled as binary MEX files. Thus, the
comparison with our library written in C++, although ``informal'', is
not completely unfounded.

Table~\ref{tab:informalcomp} gives the convex optimization
solver times (which thus do not include the robot dynamics computation
times) reported by~\cite{VerX09tac,DebX13tr} as well as the running
times of our implementation after subtracting the robot dynamics
computation times. Different grid sizes ($N$) and number of degrees of
freedom to optimize (DOF) were reported or tested.

\begin{table}[th]
  \caption{Informal comparison, torque constraints}    
  \centering
    \begin{tabular}{|l|c|c|c|c|}
      \hline
      Source&Lang.&DOF&$N$&Exec.\\
      \hline      
      Verscheure et al 2009~\cite{VerX09tac}&MEX&3&299&0.74 s\\
      Verscheure et al 2009~\cite{VerX09tac}&MEX&3&1999&2.87 s\\
      Debrouwere et al 2013~\cite{DebX13tr}&MEX&7&100&1.5 s\\
      This article&C++&3&300&0.017 s\\
      This article&C++&3&2000&0.10 s\\
      This article&C++&7&100&0.0058 s\\
      \hline
     \end{tabular}
 \label{tab:informalcomp}  
\end{table}

Next, in order to make a ``formal'' comparison -- on the same computer
and using the same programming language, we considered MINTOS
(\url{http://www.iu.edu/~motion/mintos/}, last accessed December
2013), Hauser's recent C++ implementation of the convex optimization
approach~\cite{Hau13rss}, which is, to our knowledge, the fastest
implementation currently available. To exclude the robot dynamics
computations -- which are independent of the TOPP problem and whose
execution times depend largely on the robot simulation software used,
we considered ``pure'' velocity and acceleration bounds. Note however
that, from the general formulation of Section~\ref{sec:general}, these
pure kinematic constraints involve exactly the same difficulties as
any other type of dynamic constraints.

We compiled and ran both implementations on our computer. We
considered 1000 random trajectories of dimension 7 and velocity and
acceleration bounds as in
Section~\ref{sec:compshiller}. Table~\ref{tab:formalcomp} reports the
average total execution times of the two implementations. It appears
from this controlled comparison that our implementation is between 7
and 15 times faster than MINTOS\,\footnote{Note that MINTOS includes
  an innovative ``constraints pruning'' step that significantly speeds
  up the execution with respect to previous implementations of the
  convex optimization approach. We are investigating how this idea can
  be integrated in the numerical integration approach.}.

\begin{table}[th]
  \caption{Formal comparison, kinematic constraints}    
  \centering
    \begin{tabular}{|l|c|c|c|c|}
      \hline
      Source&Lang.&DOF&$N$&Exec.\\
      \hline      
      Hauser 2013~\cite{Hau13rss}&C++&7&300&0.025 s\\
      Hauser 2013~\cite{Hau13rss}&C++&7&1000&0.20 s\\
      This article&C++&7&300&0.0038 s\\
      This article&C++&7&1000&0.014 s\\
      \hline
     \end{tabular}
 \label{tab:formalcomp}  
\end{table}

\section{Conclusion}
\label{sec:conclusion}

We have established a rigorous characterization and treatment of
dynamic singularities that arise in the numerical integration approach
to the Time-Optimal Path Parameterization (TOPP) problem. This fully
completes the celebrated line of research on TOPP, which started in the
1980's with the seminal papers of Bobrow et al.~\cite{BobX85ijrr} and
Shin and McKay~\cite{SM86tac} and which has since then received
contributions from many prominent research groups
(e.g.~\cite{PJ87jra,SY89tra,SL92jdsmc,KS12rss}, just to cite a few).

Based on that contribution and on a general formulation of the TOPP
problem, we provided an open-source implementation of TOPP. We showed
that this implementation is robust and fast: on typical test cases, it
is about one order of magnitude faster than the fastest currently
available implementation of the convex optimization approach. As our
implementation is open-source and has been designed so as to
facilitate the integration of new systems dynamics and constraints, we
hope that it will be useful to robotics researchers interested in
kinodynamic motion planning.

Our next immediate goal is to attack the \emph{feasibility} problem
(i.e. finding a collision-free, dynamically-balanced trajectory in a
challenging context, in particular, where quasi-static trajectories
are impossible) for humanoid robots using Admissible Velocity
Propagation (AVP)~\cite{PhaX13rss}, which itself is based on TOPP. We
believe that the power of AVP combined with the speed of the present
TOPP implementation can make possible the planning of unprecedentedly
dynamic motions for humanoid robots in challenging environments.

Finally, from a theoretical perspective, we are investigating how
higher-order constraints, such as jerk~\cite{TS93icra,CC00jrs} or
other types of optimization objectives~\cite{VerX09tac,DebX13tr}, can be
integrated in our TOPP framework.

\subsection*{Acknowledgments} We would like to thank St\'ephane Caron,
Zvi Shiller and Yoshihiko Nakamura for stimulating discussions
regarding TOPP. We are also grateful to S. C. and Rosen Diankov for
their helps with the implementation. This work was supported by
``Grants-in-Aid for Scientific Research'' for JSPS fellows and by a
JSPS postdoctoral fellowship.

\appendix

\subsection{Characterizing dynamic singularities}

In line with Section~\ref{sec:charact}, consider a zero-inertia point
$s^*$ and assume that $a_k(s^*)=0$ and that $a_k(s)<0$ in a
neighborhood to the left of $s^*$ and $a_k(s)>0$ in a neighborhood to
the right of $s^*$. We have the following definition and proposition.

\begin{Prop}
  \label{prop:alphabeta}
  Define, for all $(s,\dot s)$,
  \[
  \tilde{\alpha}(s,\dot s) = \max_{i\neq k} \alpha_i(s,\dot s) \ ;\ 
  \tilde{\beta}(s,\dot s) = \min_{i\neq k} \beta_i(s,\dot s).
  \]
  There exists a neighborhood $]s^*-\epsilon,s^*[$ to the \emph{left}
  of $s^*$ such that
  \[
  \forall (s,\dot s) \in ]s^*-\epsilon,s^*[\times
  [0,\infty[,\ \beta(s,\dot s) = \tilde{\beta}(s,\dot s).
  \]   
  and a neighborhood $]s^*,s^*+\epsilon[$ to the
  \emph{right} of $s^*$ such that
  \[
  \forall (s,\dot s) \in ]s^*,s^*+\epsilon[\times
  [0,\infty[,\ \alpha(s,\dot s) = \tilde{\alpha}(s,\dot s),
  \]
  Note also that $\tilde{\alpha}$ and $\tilde{\beta}$ are continuous
  and differentiable in a neighborhood \emph{around} $s^*$.
\end{Prop}

\emph{Proof\,:} From our assumption that $a_k(s)<0$ on the left of
$s^*$ and $a_k(s)>0$ on the right of $s^*$, constraint $k$ gives rise
to an $\alpha_k$ on the left of $s^*$ and to a $\beta_k$ on the right
of $s^*$. It thus does not contribute to the value of $\beta$ on the
left of $s^*$ or to that of $\alpha$ on the right of $s^*$ $\Box$

\begin{Prop}
  \label{prop:c}
  If $c_k(s^*)>0$ then there exists a neighborhood
  $]s^*-\epsilon,s^*[$ such that $\MVC(s)=0$ for all
      $s\in]s^*-\epsilon,s^*[$ (which in turn implies that the path is
      not traversable).
\end{Prop}

\emph{Proof\,:} Suppose  that $c_k(s^*)=\eta>0$.  By
continuity of $c_k$, there exists a neighborhood
$]s^*-\epsilon_1,s^*[$ such that
\[
\forall s\in ]s^*-\epsilon_1,s^*[, \ -c_k(s)<-\eta/2.
\]
On the other hand, one has $a_k(s) \uparrow 0$ when $s\uparrow
s^*$. Thus, $\alpha_k(s,0) = \frac{-c_k(s)}{a_k(s)} \to \infty$ when
$s\uparrow s^*$. Since $\alpha=\max_i\alpha_i$, we have that
$\alpha(s,0) \to +\infty$ when $s\uparrow s^*$. Next, from
Proposition~\ref{prop:alphabeta}, $\beta$ is continuous, hence
\emph{upper-bounded}, in a neighborhood to the left of $s^*$. Thus,
there exists a neighborhood to the left of $s^*$ in which
$\alpha(s,0)>\beta(s,0)$, which in turn implies that $\MVC(s)=0$ in
that neighborhood $\Box$

In light of Proposition~\ref{prop:c}, we assume from now on that
$c_k(s^*)<0$. We next distinguish two cases according to the value of
$b_k(s^*)$.


\textbf{Case $b_k(s^*)>0$:} Define
\begin{equation}
  \label{eq:sdotstar}
  \dot{s}^*=\sqrt{\frac{-c_k(s^*)}{b_k(s^*)}}.
\end{equation}
Note that, since $c_k(s^*)<0$ and $b_k(s^*)>0$, the expression under
the radical sign is indeed positive. Next, let $\dot{s}^\dag{}$ be the
smallest velocity $\dot s$ that satisfies $\tilde{\alpha}(s^*,\dot
s)=\tilde{\beta}(s^*,\dot s)$ ($\dot{s}^\dag{}=+\infty$ if no such
$\dot s$ exists).

We now distinguish two sub-cases.

\paragraph{Sub-case $\dot{s}^\dag{}<\dot s^*$}

Let $\dot s^\ddag{}=(\dot{s}^\dag{}+\dot s^*)/2$. By the definition of
$\dot{s}^*$ and the assumption that $b_k(s^*)>0$, there exists
$\eta>0$ such that, in a neighborhood to the left of~$s^*$ 
\[
\forall \dot s\leq \dot s^\ddag{},\ 
-b_k(s)\dot s^2-c_k(s)>\eta.
\]
On the other hand, one has $a_k(s) \uparrow 0$ when $s\uparrow
s^*$. Thus, $\alpha_k(s,\dot s) = \frac{-b_k(s)\dot
  s^2-c_k(s)}{a_k(s)} \to -\infty$ when $s\uparrow s^*$ and $\dot
s\leq \dot{s}^\ddag$. As a consequence, constraint $k$ does \emph{not}
contribute to $\alpha$ in a neighborhood to the left of $s^*$ and for
$\dot s \leq \dot s^\ddag{}$. Thus, one has $\alpha = \tilde{\alpha}$
in a neighborhood to the left of $s^*$ and for $\dot s \leq \dot
s^\ddag{}$.

By the same argument, one can show that $\beta = \tilde{\beta}$ in a
neighborhood to the right of $s^*$ and for $\dot s \leq \dot
s^\ddag{}$. Combined with Proposition~\ref{prop:alphabeta}, one has
thus obtained that $\alpha = \tilde{\alpha}$ and $\beta =
\tilde{\beta}$ in a neighborhood \emph{around} $s^*$ and for $\dot s
\leq \dot s^\ddag{}$. This shows that $\MVC(s^*) =\dot{s}^\dag$, and
that the $\MVC$ is entirely determined by $\tilde{\alpha}$ and
$\tilde{\beta}$ around $(s^*,\dot{s}^\dag)$. One can thus conclude
that the $\MVC$ is continuous and differentiable at $s^*$, which in
turn implies that constraint $k$ does \emph{not} trigger a singularity
at $s^*$.

\paragraph{Sub-case $\dot{s}^\dag{}>\dot s^*$}

Remark first that, excepting degenerate cases, one can find a
neighborhood $]s^*-\epsilon,s^*+\epsilon[\times ] \dot s^*-\eta, \dot
s^*+\eta[$ in which $\tilde{\alpha}$ is given by a unique $\alpha_q$
and $\tilde{\beta}$ is given by a unique $\beta_p$. Note that, by
definition of $\tilde{\alpha}$ and $\tilde{\beta}$, one has $p\neq k$
and $q\neq k$.

In the neighborhood just defined, let
\[
u(s)=\frac{-a_k(s)c_q(s) + a_q(s)c_k(s)}{a_k(s)b_q(s)-a_q(s)b_k(s)};
\]
\[
v(s)=\frac{-a_k(s)c_p(s)+a_p(s)c_k(s)}{a_k(s)b_p(s)-a_p(s)b_k(s)}.
\]

From the assumption that $a_k(s^*)=0$, one has
\[
\lim_{s\uparrow s^*} u(s)=\frac{-c_k(s^*)}{b_k(s^*)}=\dot{s}^{*2}.
\]
Thus, in a neighborhood to the left of $s^*$, one has $0<u(s)<\dot
s^{\ddag{}2}$.  Next, remark that by definition $\dot s=\sqrt{u(s)}$
satisfies $\alpha_k(s,\dot s)=\beta_q(s,\dot s)$. The above two
statements together imply that $\MVC(s) = \sqrt{u(s)}$.

One can show similarly that there exists a neighborhood to the right
of $s^*$ in which $\MVC(s)=\sqrt{v(s)}$. Combining the results
concerning the left and the right of~$s^*$, one obtains that the
$\MVC$ is \emph{continuous} at $s^*$, since
\[
\lim_{s\uparrow s^*} \MVC(s) = \lim_{s\uparrow s^*} \sqrt{u(s)} = 
\dot s^* = \lim_{s\downarrow s^*} \sqrt{v(s)} = \lim_{s\downarrow s^*} \MVC(s) . 
\]
However, the $\MVC$ is \emph{undifferentiable} at $s^*$ since, in
general,
\[
\lim_{s\uparrow s^*} \MVC'(s) = \lim_{s\uparrow s^*}
\left(\sqrt{u(s)}\right)' \neq 
\lim_{s\downarrow s^*} \left(\sqrt{v(s)}\right)' = \lim_{s\downarrow s^*} \MVC'(s) .
\]
Thus, in this sub-case, $s^*$ is indeed a dynamic singularity.


\textbf{Case $b_k(s^*)<0$:} From the assumptions that $c_k(s^*)<0$ and
$b_k(s^*)<0$, one has that $-b_k(s^*)\dot s^2-c_k(s^*)>0$ for all
$\dot s$. Thus, by the same argument as in sub-case
$\dot{s}^\dag{}<\dot s^*$, there exists a neighborhood
$]s^*-\epsilon,s^*[$ where $\alpha = \tilde{\alpha}$ and a
neighborhood $]s^*,s^*+\epsilon'[$ where $\beta = \tilde{\beta}$. One
can thus conclude that constraint $k$ does \emph{not} trigger a
singularity at $s^*$.

\bibliographystyle{IEEEtran} 
\bibliography{cri}

\begin{thebibliography}{10}
\providecommand{\url}[1]{#1}
\csname url@samestyle\endcsname
\providecommand{\newblock}{\relax}
\providecommand{\bibinfo}[2]{#2}
\providecommand{\BIBentrySTDinterwordspacing}{\spaceskip=0pt\relax}
\providecommand{\BIBentryALTinterwordstretchfactor}{4}
\providecommand{\BIBentryALTinterwordspacing}{\spaceskip=\fontdimen2\font plus
\BIBentryALTinterwordstretchfactor\fontdimen3\font minus
  \fontdimen4\font\relax}
\providecommand{\BIBforeignlanguage}[2]{{%
\expandafter\ifx\csname l@#1\endcsname\relax
\typeout{** WARNING: IEEEtran.bst: No hyphenation pattern has been}%
\typeout{** loaded for the language `#1'. Using the pattern for}%
\typeout{** the default language instead.}%
\else
\language=\csname l@#1\endcsname
\fi
#2}}
\providecommand{\BIBdecl}{\relax}
\BIBdecl

\bibitem{Lav06book}
S.~LaValle, \emph{Planning algorithms}.\hskip 1em plus 0.5em minus 0.4em\relax
  Cambridge Univ Press, 2006.

\bibitem{DonX93acm}
B.~Donald, P.~Xavier, J.~Canny, and J.~Reif, ``Kinodynamic motion planning,''
  \emph{Journal of the ACM (JACM)}, vol.~40, no.~5, pp. 1048--1066, 1993.

\bibitem{Bob88jra}
J.~Bobrow, ``Optimal robot plant planning using the minimum-time criterion,''
  \emph{IEEE Journal of Robotics and Automation}, vol.~4, no.~4, pp. 443--450,
  1988.

\bibitem{SD91tra}
Z.~Shiller and S.~Dubowsky, ``On computing the global time-optimal motions of
  robotic manipulators in the presence of obstacles,'' \emph{IEEE Transactions
  on Robotics and Automation}, vol.~7, no.~6, pp. 785--797, 1991.

\bibitem{PhaX13rss}
Q.-C. Pham, S.~Caron, and Y.~Nakamura, ``Kinodynamic planning in the
  configuration space via velocity interval propagation,'' in \emph{Robotics:
  Science and System}, 2013.

\bibitem{SM86tac}
K.~Shin and N.~McKay, ``Selection of near-minimum time geometric paths for
  robotic manipulators,'' \emph{IEEE Transactions on Automatic Control},
  vol.~31, no.~6, pp. 501--511, 1986.

\bibitem{VerX09tac}
D.~Verscheure, B.~Demeulenaere, J.~Swevers, J.~De~Schutter, and M.~Diehl,
  ``Time-optimal path tracking for robots: A convex optimization approach,''
  \emph{IEEE Transactions on Automatic Control}, vol.~54, no.~10, pp.
  2318--2327, 2009.

\bibitem{Hau13rss}
K.~Hauser, ``Fast interpolation and time-optimization on implicit contact
  submanifolds,'' in \emph{Robotics: Science and Systems}, 2013.

\bibitem{DebX13tr}
F.~Debrouwere, W.~Van~Loock, G.~Pipeleers, Q.~Tran~Dinh, M.~Diehl,
  J.~De~Schutter, and J.~Swevers, ``Time-optimal path following for robots with
  convex-concave constraints using sequential convex programming,'' \emph{IEEE
  Transactions on Robotics}, 2013.

\bibitem{SL92jdsmc}
Z.~Shiller and H.~Lu, ``Computation of path constrained time optimal motions
  with dynamic singularities,'' \emph{Journal of dynamic systems, measurement,
  and control}, vol. 114, p.~34, 1992.

\bibitem{KS12rss}
T.~Kunz and M.~Stilman, ``Time-optimal trajectory generation for path following
  with bounded acceleration and velocity,'' in \emph{Robotics: Science and
  Systems}, vol.~8, 2012, pp. 09--13.

\bibitem{Pha13iros}
Q.-C. Pham, ``Characterizing and addressing dynamic singularities in the
  time-optimal path parameterization algorithm,'' in \emph{IEEE/RSJ
  International Conference on Intelligent Robots and Systems}, 2013.

\bibitem{BobX85ijrr}
J.~Bobrow, S.~Dubowsky, and J.~Gibson, ``Time-optimal control of robotic
  manipulators along specified paths,'' \emph{The International Journal of
  Robotics Research}, vol.~4, no.~3, pp. 3--17, 1985.

\bibitem{PN12humanoids}
Q.-C. Pham and Y.~Nakamura, ``Time-optimal path parameterization for critically
  dynamic motions of humanoid robots,'' in \emph{IEEE-RAS International
  Conference on Humanoid Robots}, 2012.

\bibitem{Hau14icra}
K.~Hauser, ``Fast dynamic optimization of robot paths under actuator limits and
  frictional contact,'' in \emph{Robotics and Automation, IEEE International
  Conference on}, 2014.

\bibitem{BL01tac}
F.~Bullo and K.~M. Lynch, ``Kinematic controllability for decoupled trajectory
  planning in underactuated mechanical systems,'' \emph{IEEE Transactions on
  Robotics and Automation}, vol.~17, no.~4, pp. 402--412, 2001.

\bibitem{PJ87jra}
F.~Pfeiffer and R.~Johanni, ``A concept for manipulator trajectory planning,''
  \emph{IEEE Journal of Robotics and Automation}, vol.~3, no.~2, pp. 115--123,
  1987.

\bibitem{SY89tra}
J.~Slotine and H.~Yang, ``Improving the efficiency of time-optimal
  path-following algorithms,'' \emph{IEEE Transactions on Robotics and
  Automation}, vol.~5, no.~1, pp. 118--124, 1989.

\bibitem{ChoX05book}
H.~Choset, K.~M. Lynch, S.~Hutchinson, G.~Kantor, W.~Burgard, L.~E. Kavraki,
  and S.~Thrun, \emph{Principles of robot motion: theory, algorithms, and
  implementations}.\hskip 1em plus 0.5em minus 0.4em\relax MIT press, 2005.

\bibitem{Zla96icra}
L.~Zlajpah, ``On time optimal path control of manipulators with bounded joint
  velocities and torques,'' in \emph{IEEE International Conference on Robotics
  and Automation}, vol.~2.\hskip 1em plus 0.5em minus 0.4em\relax IEEE, 1996,
  pp. 1572--1577.

\bibitem{LL98er}
F.~Lamiraux and J.-P. Laumond, ``From paths to trajectories for multi-body
  mobile robots,'' in \emph{Experimental Robotics V}.\hskip 1em plus 0.5em
  minus 0.4em\relax Springer, 1998, pp. 301--309.

\bibitem{WO82jdsmc}
M.~Walker and D.~Orin, ``Efficient dynamic computer simulation of robotic
  mechanisms,'' \emph{Journal of Dynamic Systems, Measurement, and Control},
  vol. 104, p. 205, 1982.

\bibitem{Dia10these}
\BIBentryALTinterwordspacing
R.~Diankov, ``Automated construction of robotic manipulation programs,'' Ph.D.
  dissertation, Carnegie Mellon University, Robotics Institute, August 2010.
  [Online]. Available:
  \url{http://www.programmingvision.com/rosen_diankov_thesis.pdf}
\BIBentrySTDinterwordspacing

\bibitem{TS93icra}
M.~Tarkiainen and Z.~Shiller, ``Time optimal motions of manipulators with
  actuator dynamics,'' in \emph{IEEE International Conference on Robotics and
  Automation}.\hskip 1em plus 0.5em minus 0.4em\relax IEEE, 1993, pp. 725--730.

\bibitem{CC00jrs}
D.~Constantinescu and E.~Croft, ``Smooth and time-optimal trajectory planning
  for industrial manipulators along specified paths,'' \emph{Journal of Robotic
  Systems}, vol.~17, no.~5, pp. 233--249, 2000.

\end{thebibliography}

\end{document}